\documentclass[a4paper,fleqn]{cas-sc}
\usepackage{hyperref}
\usepackage{color}
\usepackage{soul}
\usepackage{graphicx}
\usepackage{setspace}

\usepackage[numbers]{natbib}
\usepackage{enumitem}
\usepackage{caption}
\usepackage{amsmath}
\usepackage{bm}
\usepackage{pifont}
\usepackage{float}
\usepackage{url}
\usepackage{subfigure}
\usepackage{threeparttable}
\usepackage{makecell}

\emergencystretch=\maxdimen
\def\tsc#1{\csdef{#1}{\textsc{\lowercase{#1}}\xspace}}
\tsc{WGM}
\tsc{QE}
\tsc{EP}
\tsc{PMS}
\tsc{BEC}
\tsc{DE}

\begin{document}
\setstretch{1.25}
\nocite{}
\let\WriteBookmarks\relax
\def\floatpagepagefraction{1}
\def\textpagefraction{.001}
\shortauthors{Xia P. et~al.}

\title [mode = title]{Multi-modal cross-domain mixed fusion model with dual disentanglement for fault diagnosis under unseen working conditions}                   
%
%
%
%
%
\author[1,2,3]{Pengcheng Xia}[style=chinese]
\author[1]{Yixiang Huang}[style=chinese]
\cormark[1]
\ead{huang.yixiang@sjtu.edu.cn}
\author[1]{Chengjin Qin}[style=chinese]
\author[1]{Chengliang Liu}[style=chinese]
%
%
\address[1]{State Key Laboratory of Mechanical System and Vibration, Shanghai Jiao Tong University, Shanghai 200240, PR China}
\address[2]{Singapore Institute of Manufacturing Technology (SIMTech), Agency for Science, Technology and Research (A*STAR), Singapore 636732, Singapore}
\address[3]{Al Centre of Excellence for Manufacturing (AIMfg), Agency for Science, Technology and Research (A*STAR), Singapore 637143, Singapore}
%
%
%
%
%
%
%
%

\begin{abstract}
Intelligent fault diagnosis has become an indispensable technique for ensuring machinery reliability. However, existing methods suffer significant performance decline in real-world scenarios where models are tested under previously unseen working conditions, while domain adaptation approaches are limited by their dependence on target-domain samples. In addition, most existing studies rely on single-modal sensing signals, overlooking the complementary nature of multi-modal information for improving model generalization. To address these limitations, this paper proposes a multi-modal cross-domain mixed fusion model with dual disentanglement for fault diagnosis. A dual disentanglement framework is developed to decouple modality-invariant and modality-specific features, as well as domain-invariant and domain-specific representations, enabling both comprehensive multi-modal representation learning and enhanced generalization to working conditions. A cross-domain mixed fusion strategy is designed to increase modality and domain diversity by randomly mixing modality information across domains. Furthermore, a triple-modal fusion mechanism is introduced to adaptively integrate multi-modal heterogeneous information. Extensive experiments are conducted on induction motor fault diagnosis under both unseen constant and time-varying working conditions. The results demonstrate that the proposed method consistently outperforms advanced methods, and comprehensive ablation studies further verify the effectiveness of each proposed component and the advantage of multi-modal fusion. The code is available at: https://github.com/xiapc1996/MMDG.
\end{abstract}



\begin{keywords}
	Fault diagnosis \sep
	Multi-modal fusion \sep
	Domain generalization \sep
	Feature disentanglement \sep
	Unseen working conditions	
\end{keywords}

\ExplSyntaxOn
\keys_set:nn { stm / mktitle } { nologo }
\ExplSyntaxOff

\maketitle

\setstretch{1.2}
\section{Introduction}
As modern machinery advances toward higher levels of automation and complexity, increasingly stringent requirements have been imposed on equipment reliability. Unexpected failures can lead to substantial economic losses, unplanned downtime, and even catastrophic accidents~\cite{review1}. To mitigate these risks, large-scale sensor networks have been widely deployed in industrial systems to continuously monitor diverse signals, such as vibration, current, and acoustic signals~\cite{multisensor}. By exploiting these monitoring data, fault diagnosis techniques enable early detection of anomalies and facilitate timely maintenance actions, thereby improving operational safety and system reliability~\cite{review2}.

Recently, intelligent data-driven fault diagnosis methods based on deep learning have attracted substantial attention owing to their strong capability to automatically learn discriminative representations from raw sensor measurements. Compared with traditional physics-based approaches and conventional machine-learning methods that rely heavily on hand-crafted feature engineering, deep learning techniques alleviate the dependence on complex prior expert knowledge and manual feature extraction, thereby demonstrating promising diagnostic performance~\cite{review3,motor}. In particular, vibration signals of rotating machinery can be directly fed into deep learning models to automatically learn fault-related features. Borghesani et al.~\cite{1DCNN} employed a one-dimensional (1D) convolutional neural network (CNN) for bearing fault diagnosis and attempted to interpret the learned vibration features. Liu et al.~\cite{MKCNN} designed a residual network with multiscale convolutional kernels for motor fault diagnosis based on vibration signals. However, raw one-dimensional vibration signals often exhibit complex temporal dynamics and non-stationary characteristics, which may limit the ability of conventional 1D models to fully capture informative patterns. To better characterize the underlying time–frequency structures, many studies transform 1D vibration signals into two-dimensional time–frequency representations using short-time Fourier transform (STFT)~\cite{STFT} or wavelet transform~\cite{time-frequency}, and subsequently employ two-dimensional networks to extract more expressive features.

Nevertheless, vibration signals may exhibit limited sensitivity to certain electrical faults in electromechanical systems such as motors~\cite{motor_review}. As a result, motor current signals have also been widely investigated for fault diagnosis~\cite{motor_DT}. For instance, Jimenez-Guarneros et al.~\cite{MODWT} proposed a lightweight 1D CNN to diagnose both mechanical and electrical faults of induction motors using current signals. Furthermore, acoustic signals also contain rich fault-related information and have attracted increasing interest for machinery fault diagnosis. Zhang et al.~\cite{acoustic_bearing} employed acoustic measurements together with a graph convolutional network to diagnose bearing faults, while Xiao et al.~\cite{acoustic_motor} utilized a denoising autoencoder to achieve motor fault diagnosis based on acoustic signals. These studies collectively demonstrate the effectiveness of deep learning-based fault diagnosis using acoustic information.

Despite the remarkable progress achieved by intelligent fault diagnosis methods, their generalization capability remains a critical challenge for practical applications, as industrial machines typically operate under variable working conditions. Domain shifts induced by changes in operating speed and load can significantly degrade the performance of models trained on data collected under specific source conditions. To alleviate this issue, domain adaptation (DA) techniques have been extensively investigated to enhance the robustness of diagnostic models under distribution shifts. By aligning feature distributions between source and target domains, these methods aim to mitigate the adverse effects of condition variability and enable more reliable cross-condition fault diagnosis~\cite{TL2}. Consequently, DA approaches based on discrepancy minimization~\cite{TL3}, domain adversarial learning~\cite{TL4}, and subdomain adaptation (SDA)~\cite{SDA} have demonstrated effectiveness in machinery fault diagnosis across varying working conditions.

Nevertheless, a major limitation that restricts the practical applicability of DA methods lies in their reliance on target-domain data for distribution alignment. In real industrial environments, this requirement is often difficult to satisfy, as machines frequently operate under newly emerging or previously unseen working conditions for which no labeled or unlabeled samples are available in advance. To overcome this constraint, domain generalization (DG) has emerged as a promising alternative, aiming to learn models that can generalize to unseen target domains without accessing any target-domain data during training~\cite{DG_review}. Rather than explicitly aligning source and target distributions, DG focuses on extracting domain-invariant and discriminative representations from multiple source domains, thereby improving robustness to distribution shifts and demonstrating encouraging performance in unseen-condition fault diagnosis~\cite{DG_review2}. Most existing DG-based fault diagnosis methods attempt to achieve this goal through domain adversarial learning~\cite{DG2} or domain divergence minimization~\cite{DG1}, so that the learned invariant features are more likely to generalize to unseen domains. In addition, several studies have explored data augmentation strategies~\cite{Augmentation} or meta-learning frameworks~\cite{Meta-learning} to further enhance generalization capability. However, the majority of existing methods primarily emphasize learning domain-invariant representations, while largely neglecting domain-specific features that are closely correlated with distinct working conditions.

Moreover, most existing DG-based fault diagnosis studies rely on a single sensing modality, typically vibration signals. However, using only one type of signal often limits diagnostic generalization, as different sensing modalities capture complementary fault-related characteristics, particularly in electromechanical systems. Consequently, in the context of DG-based fault diagnosis, multi-modal fusion provides additional potential for improving robustness under unseen operating conditions. By integrating information from multiple modalities, the diagnostic model can reduce its dependence on any single domain-specific feature distribution and achieve more stable cross-domain generalization. Although numerous studies have investigated multi-sensor fusion strategies for fault diagnosis~\cite{fusion}, including data-level fusion~\cite{data_fusion}, feature-level fusion~\cite{feature_fusion}, and decision-level fusion~\cite{multisensor}, significant challenges remain when operating conditions vary. First, most existing approaches are designed to process one-dimensional signals from multiple sensors with identical sampling frequencies, which limits their applicability when time-frequency representations are employed or when different modalities inherently operate at different sampling rates. Second, many fusion methods directly combine data or features from multiple sensors without explicitly modeling the intrinsic correlations among modalities or the modality-specific characteristics associated with different fault types. Furthermore, to the best of our knowledge, fault diagnosis under unseen working conditions using multi-modal data has not yet been adequately addressed in the literature. As a result, the design of an effective multi-modal fusion framework that can enhance diagnostic generalization to unseen operating conditions remains a challenging and open research problem.

To tackle the aforementioned challenges and limitations, this paper proposes a multi-modal cross-domain mixed fusion model with dual disentanglement for fault diagnosis. First, multi-modal data are encoded by respective encoders to obtain dedicated modality embeddings. Second, a cross-domain mixed fusion strategy is proposed to randomly mix each modality across various source domains, thereby mitigating domain bias and enriching cross-domain feature diversity. Subsequently, a dual disentanglement framework is designed to separately disentangle modality-invariant and modality-specific representation, and domain-invariant and domain-specific representation disentanglement, enabling more robust domain generalization. Furthermore, a triple-modal fusion module based on multiple cross-attention is developed to achieve deep and adaptive fusion of heterogeneous modalities. To evaluate the proposed method, we conducted experiments on induction motors and collected vibration, current, and acoustic signals under both constant and varying working conditions. Extensive fault diagnosis experiments on unseen working conditions were performed, and both comparative and ablation studies consistently validate the effectiveness and superiority of the proposed method. The main contributions of this work are summarized as follows:

\begin{enumerate}[leftmargin=2\parindent, noitemsep]
	\item[(1)] A dual disentanglement framework jointly disentangling invariance and specificity at both the modality level and the domain level is proposed for enhanced multi-modal DG for fault diagnosis.
	\item[(2)] A multi-modal cross-domain mixed fusion mechanism is designed for modality augmentation to mitigate domain bias and enhance generalization.
	\item[(3)] A triple-modal fusion module is introduced to achieve adaptive and complementary fusion of multiple heterogeneous modalities.
\end{enumerate}

The remainder of the paper is organized as follows. Section~\ref{sec_work} introduces some related work regarding multi-modal fusion and DG. The proposed method is described in detail in Section~\ref{sec_method}. The experiments and results are presented in Section~\ref{sec_experiment}, and Section~\ref{sec_conclusion} concludes the paper.

\section{Related work}\label{sec_work}
\subsection{Domain generalization}
Different from DA, DG removes the constraint of access to target domain data during training. This setting is more consistent with real-world fault diagnosis scenarios, where data from new working conditions are typically unavailable in advance. As a result, DG has attracted extensive research attention~\cite{DG_review3}. The mainstream idea in DG is to learn invariant representations across multiple source domains. Therefore, domain adversarial learning has become the most widely adopted approach, which introduces a domain discriminator to adversarially train the feature extractor to learn domain-invariant features that are able to confuse the discriminator. For example, Chen et al.~\cite{adversarial1} leveraged adversarial learning to exploit domain-invariant features for bearing fault diagnosis under unseen working conditions. Shi et al.~\cite{adversarial2} designed a weighting strategy for domain adversarial learning based on domain transferability, achieving remarkable performance on multiple bearing datasets. Some works explicitly aligned feature distributions from multiple source domains. For example, Pu et al.~\cite{DG1} introduced $\alpha$-PE divergence for distribution alignment for gearbox fault diagnosis under variable working conditions. 

However, these methods primarily emphasize the extraction of domain-invariant features while neglecting the importance of domain-specific characteristics that are closely associated with individual working conditions. This leads to the loss of condition-related information, which can limit both robustness and classification performance when the model encounters working conditions differing substantially from those during training. Consequently, Zhao et al.~\cite{specificity} proposed to combine invariance and specificity for fault diagnosis generalization and proposed a DG network with separate subnetworks for domain-invariant and domain-specific feature extraction, and the similarity to source domains is estimated to select specificity for each target sample. Other studies have attempted to address this issue from the perspective of causal learning by simultaneously modeling causal and non-causal features. Li et al.~\cite{CCN} proposed a causal consistency network (CCN) to learn causal features with machines and conditions jointly. Similarly, Jia et al.~\cite{CDDG} proposed to disentangle fault-related causal factors and domain-related non-causal factors by a causal aggregation loss. He et al.~\cite{motor_causal} extracted the non-causal factors with a domain classifier and decoupled them with causal features by a mutual information loss for induction motor fault diagnosis. Although these approaches improve generalization, they mostly involve additional classifiers or decoders for causal and non-causal factor learning, or incorporate complex modules for specificity extraction and selection. Moreover, all existing methods were designed for single-modal data.

In addition, several studies have introduced domain augmentation strategies to improve generalization across domains. For example, Fan et al.~\cite{DMDGN} adopted Mixup-based data augmentation~\cite{mixup} in both class and domain spaces to generate new samples for enhanced DG. Zhao et al.~\cite{AMIG} proposed a domain generation module to synthesize new domain samples, thereby improving DG performance when only limited source-domain samples are available. Nevertheless, these data generation/augmentation-based DG methods are mainly designed for single-modal sample generation. As a result, they cannot explicitly model the cross-domain diversity of different sensing modalities, and their benefits may be limited in multi-modal DG scenarios where modality heterogeneity and domain shifts coexist.

\subsection{Multi-modal fusion}
In the context of fault diagnosis, fusing signals from multiple sensors has received substantial attention. The fusion strategies can generally be categorized into three types: data-level fusion, feature-level fusion, and decision-level fusion~\cite{multisensor}. Data-level fusion directly combines raw data from multiple sensors without any information loss~\cite{data_fusion2}, while it is often limited to single-modal or homogeneous data. Feature-level fusion integrates intermediate representations for complementary feature combination from multiple signals. Some studies fused multi-source representations through direct feature concatenation~\cite{feature_fusion2}, while others rely on average pooling operations~\cite{feature_fusion3}. More recently, cross-attention mechanisms have been introduced to enhance multi-feature interaction~\cite{feature_fusion4}, though existing work focuses on fusing only two feature sources, limiting its scalability. Decision-level fusion aggregates the outputs of multiple classifiers or decision modules through decision theory such as Dempster-Shafer (D-S) evidence theory~\cite{MFHR} and soft-voting rule~\cite{decision_fusion2}, while these methods rely heavily on the accuracy of individual models.

Some studies have considered information fusion from multiple heterogeneous modalities for fault diagnosis. For example, Sun et al.~\cite{multimodal1} fused acoustic signals and infrared thermal (IRT) images with a correlation fusion module for gearbox and bearing-rotor system fault diagnosis. Ying et al.~\cite{multimodal2} also proposed a feature fusion module based on the Dirichlet distribution and D-S evidence theory to fuse acoustic signals and IRT images for fault diagnosis. However, these methods mainly focus on information fusion while overlooking the specificity within each modality, and they were designed for constant working conditions without condition bias. Recently, Liu et al.~\cite{MMGAN} proposed a multi-modal feature collaborative learning method, which learned modal-shared and modal-specific features collaboratively for compound fault detection. Despite that intra-modal specificity was taken into consideration, the domain shift issue was not investigated based on the multi-modal information. Zhang et al.~\cite{MMDA} investigated a DA method with multi-modal fusion by combining MMD loss and modality consistency loss. Their approach enables fault diagnosis across working conditions, while it follows a DA paradigm rather than DG, and thus requires access to target-domain data. Similarly, Lu et al.~\cite{MMTL} established a transfer learning method with multi-modal sensors for tool life prediction. Yu et al.~\cite{MMFSL} designed a multi-modal fusion-based method for rotating machinery fault diagnosis, which integrated meta-learning for cross-condition scenarios with a small amount of data. However, these methods rely on the availability of target-domain samples, either for training or as inference support samples, which is less practical than our setting where no data is available from the unseen target condition.

\subsection{Methodological comparison}
To explicitly compare the proposed method with related works, a methodological comparison table is summarized as Table~\ref{tab:related_methods}. Existing DG-based fault diagnosis methods mainly focus on learning domain-invariant representations from single-modal signals, and only a few studies preserve domain-specific information for generalization. Data generation or augmentation methods introduce sample generation for DG, while cannot enhance modality-level robustness by explicitly modeling diversity of each modality. In contrast, most multi-modal fault diagnosis methods emphasize information fusion but do not explicitly address unseen working conditions even though some works involve DA setting. Although multi-modal disentanglement has been studied in general multi-modal learning, they neglect the modality-level variance caused by domain shift. Therefore, the proposed method is distinguished from existing studies by jointly modeling two types of heterogeneity of sensing-modality and working-condition. Under this newly developed dual-disentanglement framework, a modality-level cross-domain augmentation strategy is proposed to promote both modality-level robustness learning and domain-level generalization learning. Furthermore, a pairwise cross-attentional fusion module is developed for triple-domain fusion setting, allowing the model to adaptively capture fine-grained inter-modal interactions.
\begin{table}[width=\linewidth,cols=6]
	\caption{Methodological comparison between the proposed method and related representative fault diagnosis studies.}\label{tab:related_methods}
	\setlength{\tabcolsep}{6pt}
	\begin{tabular*}{\tblwidth}{cccccc}
		\toprule
		Method & DG setting & \makecell{Multi-modal\\data fusion} & \makecell{Modality-level\\disentanglement} & \makecell{Domain-level\\disentanglement} & \makecell{Modality-level\\generation} \\
		\midrule
		\makecell{Invariance learning-based DG\\\cite{DG1}, \cite{adversarial1}, \cite{adversarial2}} & \checkmark & \ding{55} & \ding{55} & -- & \ding{55} \\
		\makecell{Causal/disentanglement DG\\\cite{specificity}, \cite{CCN}, \cite{CDDG}, \cite{motor_causal}} & \checkmark & \ding{55} & \ding{55} & \checkmark & \ding{55} \\
		\makecell{Generation/augmentation-based DG\\\cite{DMDGN}, \cite{AMIG}} & \checkmark & \ding{55} & \ding{55} & -- & \makecell{--\\(Sample-level)} \\
		\makecell{Multi-modal fusion \\\cite{multimodal1}, \cite{multimodal2}, \cite{MMGAN}} & \ding{55} & \makecell{\checkmark\\(2 modalities)} & -- & \ding{55} & \ding{55} \\
		\makecell{Multi-modal DA\\\cite{MMDA}, \cite{MMTL}, \cite{MMFSL}} & \ding{55} & \makecell{\checkmark\\(2 modalities)} & -- & -- & \ding{55}\\
		Proposed method & \checkmark & \makecell{\checkmark\\(3 modalities)} & \checkmark & \checkmark & \checkmark \\
		\bottomrule
	\end{tabular*}
	\begin{tablenotes}
		\item ``--'' indicates that the method partially involves this aspect, but only in an indirect, limited, or different setting from that considered in this work.
	\end{tablenotes}
\end{table}

\section{Methodology}\label{sec_method}
\subsection{Overview of the method}\label{sec_overview}
In this paper, we aim to address the multi-modal domain generalization problem for fault diagnosis under unseen working conditions. We assume there exist $M$ source domains $\{\mathscr{D}_m^s\}_{m=1}^M$, each corresponding to a specific working condition. In the $m$-th source domain, there are $N_m^s$ samples which all have fault labels, forming a sub-dataset $D_m^s=\left\{(\mathbf{x}_i^{s_m}, y_i^{s_m})\right\}_{i=1}^{N_m^s}$. Multichannel vibration, current, and acoustic signals collected from multiple source domains are used in this work. Therefore, each sample $\mathbf{x}_i^{s_m}$ consists of three heterogeneous modalities, which can be expressed as $\mathbf{x}_i^{s_m}=\left(\mathbf{x}_i^{s_m,v}, \mathbf{x}_i^{s_m,c}, \mathbf{x}_i^{s_m,a}\right)$, where $\mathbf{x}_i^{s_m,v}\in\mathcal{X}^{(v)}$, $\mathbf{x}_i^{s_m,c}\in\mathcal{X}^{(c)}$, $\mathbf{x}_i^{s_m,a}\in\mathcal{X}^{(a)}$. Here, $\mathcal{X}^{(v)}\subset \mathbb{R}^{L_v\times C_v}$ denotes the space of $C_v$ channel vibration signals with a length of $L_v$. $\mathcal{X}^{(c)}\subset \mathbb{R}^{L_c\times C_c}$ represents the space of $C_c$ channel current signals with a length of $L_c$. And $\mathcal{X}^{(a)}\subset \mathbb{R}^{L_a\times C_a}$ represents the space of $C_a$ channel acoustic signals with a length of $L_a$. Accordingly, the overall multi-modal heterogeneous input space is formulated as the Cartesian product, i.e., $\mathbf{x}_i^{s_m}\in \mathcal{X}=\mathcal{X}^{(v)}\times\mathcal{X}^{(c)}\times\mathcal{X}^{(a)}$. All the source domains share a common fault label space $\mathcal{Y}=\{1,2,\ldots,K\}$, i.e., $y_i^{s_m}\in\mathcal{Y}$ for each source sample, where $K$ is the number of health condition class. The complete multi-domain training data are obtained as $D^s=\bigcup_{m=1}^MD_m^s$. As the source domain data comes from multiple working conditions, the marginal probability distributions are non-identical among different domains, i.e., $P_i^s(X)\neq P_j^s(X)$, $i\neq j$, where $\mathscr{D}_m^s=\{\mathcal{X}_m^s,P_m^s(X)\}$. The objective of this work is to learn a model $f:\mathcal{X}\rightarrow \mathcal{Y}$ that generalizes to the previously unseen target domain $\mathscr{D}^t=\{\mathcal{X}^t,P^t(X)\}$ without accessing any data from this domain. Notably, $P^t(X)\neq P_m^s(X)$, $\forall m=1,2,\ldots,M $, while the label space remains consistent across domains, i.e., $\mathcal{Y}^t=\mathcal{Y}$.

In this problem, the observed signals are jointly influenced by multiple underlying factors, including the fault type, the sensing modality, and the operating condition. Specifically, the fault type determines the intrinsic fault-related patterns, while the sensing modality governs how these patterns are captured by different sensors (e.g., vibration, current, and acoustic), and the operating condition further modulates these manifestations under varying speeds and loads. Let $y$ denote the fault label, $V_M$ the modality variable, and $V_D$ the domain variable corresponding to different working conditions. The observed signal $\mathbf{x}$ can be modeled as a function of these factors:
\begin{equation}
	\mathbf{x}=\varPhi(y,V_M,V_D,\epsilon),
\end{equation}
where $\varPhi(\cdot)$ denotes the underlying data generation process and $\epsilon$ represents stochastic noise. From a representation learning perspective, this process can be further factorized into latent components:
\begin{equation}
	Z_f=f(y),Z_m=g(V_M),Z_d=h(V_D),
\end{equation}
\vspace{-1cm}
\begin{equation}
	\mathbf{x}=\varPsi(Z_f,Z_m,Z_d),
\end{equation}
where $Z_f$ captures fault-related intrinsic characteristics, $Z_m$ represents modality-dependent observation patterns, and $Z_d$ encodes domain-specific variations induced by operating conditions. $\varPsi(\cdot)$ denotes an abstract composition function that integrates these latent factors to generate the observed signals.

Under this view, the sensing modality and the operating domain are assumed to act as two independent sources of variation, rather than forming a direct causal relationship with each other. Instead, they jointly affect the observed signals through different mechanisms. Modality heterogeneity arises from different sensing mechanisms and directly affects the feature space of the input signals, leading to discrepancies in the representation space. In contrast, domain shift mainly manifests as distribution variation within a given representation space due to changing operating conditions. Consequently, we propose a progressive disentanglement strategy, where modality-level disentanglement is first applied to decouple variations induced by heterogeneous sensing mechanisms and to extract cross-modal shared representations and specific representations. Subsequently, the domain-level disentanglement is performed on the fused representations to further decouple variations caused by different operating conditions.

\begin{figure}
	\centering
	\includegraphics[width=\linewidth]{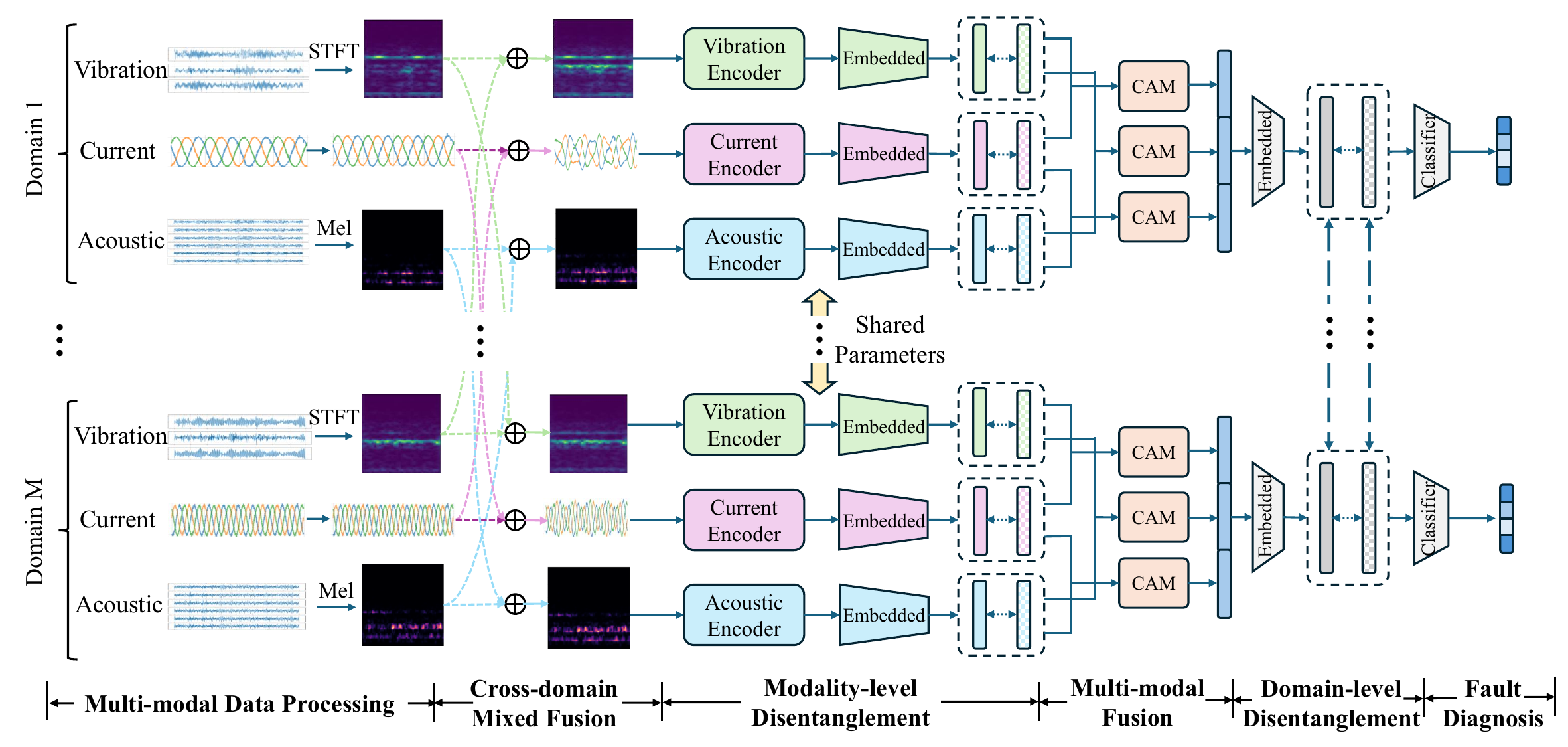}
	\caption{Framework illustration of the proposed method. STFT: Short-time Fourier transform. CAM: Cross-attention mechanism.}
	\label{fig:framework}
\end{figure}

Based on this design, a multi-modal cross-domain mixed fusion model with dual disentanglement is proposed for fault diagnosis under unseen working conditions. The framework is illustrated as Fig.~\ref{fig:framework}. Preprocessing is first conducted, including transforming vibration signals into time-frequency representations and transforming acoustic signals into Mel-spectrograms. Cross-domain mixed fusion is subsequently applied to randomly mix each modality across source domains to achieve modality augmentation. Afterward, modality-invariant representations shared by the three modalities and modality-specific representations unique to each modality are extracted and disentangled. These disentangled features from three modalities are subsequently fused through a triple-modal fusion module to form comprehensive representations for each source domain. It is worth noting that the modality-specific representations are preserved for information fusion as different sensing modalities capture complementary aspects of the underlying system dynamics and modality-specific features can serve as important complementary information for diagnosis. Furthermore, domain shifts induced by varying working conditions do not affect all sensing modalities in a uniform manner. Instead, certain fault-related cues may remain more stable or more discriminative in one modality than in others under domain shift. Therefore, retaining modality-specific components allows the model to preserve complementary information that is beneficial for potential robustness to distinguish fault categories under unseen working conditions. Subsequently, domain-invariant and domain-specific features are further learned and disentangled across all source domains to capture both shared and condition-dependent characteristics. Finally, a fault classifier is attached to accomplish fault diagnosis.

\subsection{Cross-domain mixed fusion}
As discussed above, 1D vibration signals often exhibit pronounced non-stationary characteristics under varying working conditions. Time-frequency representations are therefore more effective for capturing fault-related patterns that evolve over time and across frequency bands. We leverage STFT to convert each vibration channel into a 2D time-frequency image, and the images from multiple channels are stacked to form a multichannel image $\hat{\mathbf{x}}_i^{s_m,v}\in\mathbb{R}^{H_v\times W_v\times C_v}$. For acoustic signals, Mel-spectrograms are utilized to project the linear frequency spectrum into a perceptually inspired nonlinear scale, which can reduce redundancy in high-frequency regions, emphasize lower-frequency components where fault information is typically concentrated, and enhance the signal-to-noise ratio. Similarly, the Mel-spectrograms from multiple channels are stacked, forming a multichannel image $\hat{\mathbf{x}}_i^{s_m,a}\in\mathcal{R}^{H_a\times W_a\times C_a}$. The current signals are directly input to the network as $\hat{\mathbf{x}}_i^{s_m,c}=\mathbb{x}_i^{s_m,c}$.

It should be noted that the network inputs of different modalities can exhibit distinct sizes as multiple unshared encoders are adopted. However, the signals from different modalities should come from identical timestamps with the same time duration to ensure temporal alignment at physical time-interval level. Consequently, the multi-modal inputs describe the same underlying machine state, even though the number of sampling points may differ across modalities due to different sampling rates. Furthermore, the proposed multi-modal fusion method does not rely on strict frame-level synchronization across modalities as the fusion is conducted at the intermediate stage at the feature representation level. As a result, strict synchronization of individual time-frequency frames is not required at the data processing stage.

The proposed cross-domain mixed fusion mechanism aims to improve robustness to domain shifts by introducing random cross-domain perturbations at the modality level. For each sample, each modality is randomly mixed with the corresponding modality with the same fault category drawn from other source domains. This operation increases intra-class modality diversity and encourages the model to learn more modality-robust and domain-agnostic representations. Specifically, for each sample $\hat{\mathbf{x}}_i^{s_m}=\left(\hat{\mathbf{x}}_i^{s_m,v}, \hat{\mathbf{x}}_i^{s_m,c}, \hat{\mathbf{x}}_i^{s_m,a}\right)$ with fault label $k$ from the $m$-th domain, data of each modality $\hat{\mathbf{x}}_i^{s_m,l}, l=\{v,c,a\}$ are performed cross-domain fusion separately. For $\hat{\mathbf{x}}_i^{s_m,l}$, there is 50\% probability to randomly choose a source domain $\mathscr{D}_n^s, n\neq m$ from the rest $M-1$ source domains, and subsequently randomly select a sample with the same label $k$, i.e., $\hat{\mathbf{x}}_j^{s_n}=\left(\hat{\mathbf{x}}_j^{s_n,v}, \hat{\mathbf{x}}_j^{s_n,c}, \hat{\mathbf{x}}_j^{s_n,a}\right)$. The mixed fusion operation is performed through linear fusion of the corresponding modality of these two samples, expressed as follows:
\begin{equation}
	\tilde{\mathbf{x}}_i^{s_m,l} = \begin{cases}
	\alpha\hat{\mathbf{x}}_i^{s_m,l} + (1-\alpha)\hat{\mathbf{x}}_j^{s_n,l}, & \beta_1<0.5\\
	\hat{\mathbf{x}}_i^{s_m,l}, & \beta_1\geq0.5
	\end{cases}, \beta_1\sim U(0,1).
\end{equation}
The fusion coefficient $\alpha$ is sampled from \textit{Beta} distribution similar to Mixup~\cite{mixup}, whereas an extra operation is introduced to realize extrapolation close to the original data. Unlike standard Mixup, which adopts symmetric interpolation to fuse different samples and labels, we introduce a constrained extrapolation to further expand the data space along the domain variation direction. The extrapolation ensures that the generated sample remains close to the original sample by constraining the coefficient close to 1 while introducing controlled domain variation. This design can preserve consistency with the original samples and produce physically meaningful perturbations. Firstly, a value $\alpha^\prime$ is obtained from \textit{Beta} distribution as
\begin{equation}
	\alpha^\prime\sim Beta(0.2, 0.2).
\end{equation}
Subsequently, extrapolation is realized with 50\% probability by generating a coefficient larger than but close to 1, formulated as
\begin{equation}
	\alpha = \begin{cases}
		2-\text{max}(\alpha^\prime, 1-\alpha^\prime), & \beta_2<0.5\\
		\alpha^\prime, & \beta_2\geq0.5
	\end{cases}, \beta_2\sim U(0,1).
\end{equation}

For all modalities in a sample, the procedure is performed separately, thereby enhancing intra-class and inter-domain modality diversity for a more robust input space. The fused output $\tilde{\mathbf{x}}_i^{s_m}=\left(\tilde{\mathbf{x}}_i^{s_m,v}, \tilde{\mathbf{x}}_i^{s_m,c}, \tilde{\mathbf{x}}_i^{s_m,a}\right)$ is subsequently input to the model.

\subsection{Modality-level disentanglement}
\begin{figure}
	\centering
	\includegraphics[width=0.8\linewidth,trim=100 50 100 20,clip]{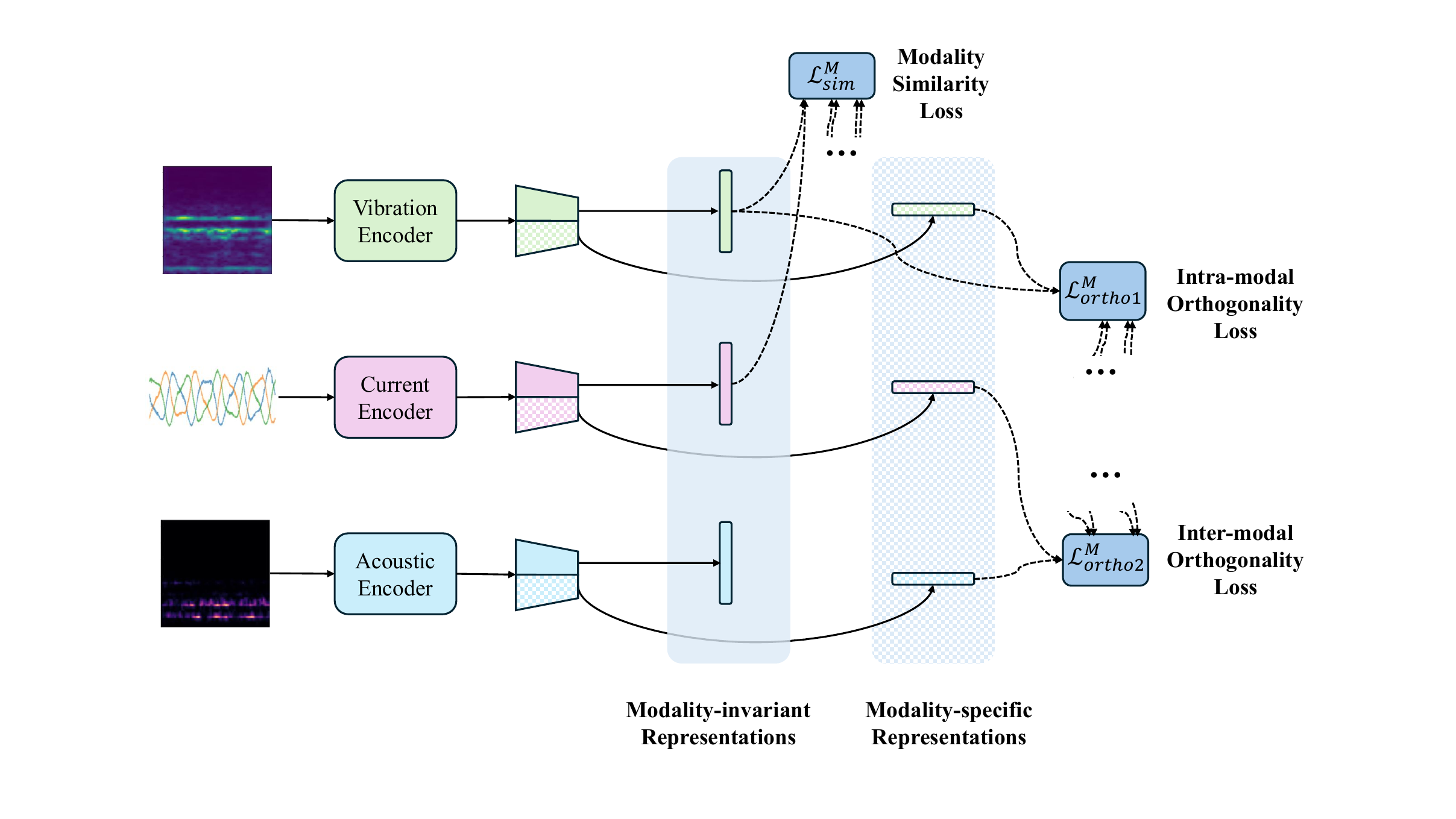}
	\caption{Illustration of modality-level disentanglement.}
	\label{fig:modality-level}
\end{figure}

Fig.~\ref{fig:modality-level} illustrates the modality-level disentanglement process. Three encoders are designed to extract representations from each sensing modality independently, implemented using residual networks (ResNets). Specifically, 2D ResNets are adopted for vibration and acoustic modalities to process their corresponding time-frequency and Mel-spectrogram images, while a 1D ResNet is employed for the current modality to handle multichannel temporal signals. Given a cross-domain mix fused modality input $\tilde{\mathbf{x}}_i^{s_m,l}$, the feature extraction can be expressed as
\begin{equation}
	\mathbf{f}_i^{s_m,l}=E^{(l)}\left(\tilde{\mathbf{x}}_i^{s_m,l}\right),l\in\{v,c,a\},
\end{equation}
where $E^{(l)}(\cdot)$ denotes the encoder for the $l$ modality.

Although these representations are extracted independently, they inherently contain both information shared across modalities and modality-specific characteristics. As multi-modal signals are collected from the same machine under identical health states and working condition, they should exhibit invariant characteristics related to the underlying fault and working conditions, thereby sharing common motives for fault diagnosis. Consequently, extracting these modality-invariant representations helps reveal correlated fault-related features consistently manifested in multiple modalities. On the other hand, each modality also contains specific features, as sensors with various modalities capture diverse aspects of machine operation. Features from different modalities may also exhibit various sensitivities to particular fault types. Therefore, aligning the modality-invariant parts while separating the modality-specific components is thus beneficial for multi-modal collaborative fault diagnosis. Inspired by~\cite{MISA}, a modality-level disentanglement strategy is introduced to explicitly decouple these two types of information.

For each modality, the extracted feature $\mathbf{f}_i^{s_m,l}$ is further transformed by two separate embedding networks, which are respectively designed to learn modality-invariant representations and modality-specific representations. Specifically, this process is formulated as
\begin{equation}
	\mathbf{z}_i^{s_m,l,inv}=G_{inv}^{(l)}\left(\mathbf{f}_i^{s_m,l}\right),\mathbf{z}_i^{s_m,l,spe}=G_{spe}^{(l)}\left(\mathbf{f}_i^{s_m,l}\right),
\end{equation}
where $G_{inv}^{(l)}(\cdot)$ and $G_{spe}^{(l)}(\cdot)$ denote the modality-invariant and modality-specific embedding networks for the $l$ modality, respectively. The modality-invariant representations $\mathbf{z}_i^{s_m,l,inv}$ are expected to encode fault-related semantics that are consistent across multiple modalities, whereas the modality-specific representations $\mathbf{z}_i^{s_m,l,spe}$ preserve distinctive characteristics arising from heterogeneous sensing mechanisms. To explicitly disentangle the modality-invariant and modality-specific components, three sets of loss functions are introduced as follows.

\textbf{(1) Modality similarity loss.} The modality-invariant representations extracted from different modalities are expected to share similar distributions in the embedding space. Consequently, a modality-level MMD loss is introduced to explicitly align the distributions of invariant representations across modalities. Specifically, the MMD loss is computed pairwise among vibration, current, and acoustic modalities within the same source domain, and the overall invariant alignment loss is defined as
\begin{equation}
	\begin{aligned}
	\mathcal{L}_{sim}^M &= \sum_{l_1\neq l_2}\text{MMD}\left(\mathbf{z}^{s_m,l_1,inv}, \mathbf{z}^{s_m,l_2,inv}\right)\\
	&=\sum_{l_1\neq l_2}\left[\frac{1}{(N_m^s)^2}\sum_{i=1}^{N_m^s}\sum_{j=1}^{N_m^s}k\left(\mathbf{z}_i^{s_m,l_1,inv}, \mathbf{z}_j^{s_m,l_1,inv}\right) + \frac{1}{(N_m^s)^2}\sum_{i=1}^{N_m^s}\sum_{j=1}^{N_m^s}k\left(\mathbf{z}_i^{s_m,l_2,inv}, \mathbf{z}_j^{s_m,l_2,inv}\right)\right.\\
	&\phantom{=\;\;}\left.- \frac{2}{(N_m^s)^2}\sum_{i=1}^{N_m^s}\sum_{j=1}^{N_m^s}k\left(\mathbf{z}_i^{s_m,l_1,inv}, \mathbf{z}_j^{s_m,l_2,inv}\right)\right],
	\end{aligned}
\end{equation}
where $k(\cdot, \cdot)$ denotes kernel function. By minimizing this loss, the invariant subspaces of different modalities are encouraged to follow similar distributions.

\textbf{(2) Intra-modality orthogonality loss.} While invariant representations should capture shared semantics, modality-specific representations are expected to encode complementary information that is independent of the invariant components. To prevent information redundancy and leakage between these two subspaces, an orthogonality constraint is imposed within each modality. Specifically, a covariance-based orthogonality loss is adopted to minimize the statistical dependency between modality-invariant and modality-specific representations. For each modality $l$, this loss is formulated as
\begin{equation}
	\mathcal{L}_{ortho1}^M = \sum_{l\in\{v,c,a\}}\left\|\text{Cov}\left(\mathbf{z}^{s_m,l,inv}, \mathbf{z}^{s_m,l,spe}\right)\right\|_2,
\end{equation}
where $\text{Cov}(\cdot)$ denotes the sample covariance matrix computed after mean normalization. Minimizing this constraint encourages the modality-specific representations to capture information that is uncorrelated with the shared invariant features.

\textbf{(3) Inter-modality orthogonality loss.} In addition to the separation within each modality, modality-specific representations from different modalities should also remain distinctive because the shared components should be included in the modality-invariant representations of each modality. Excessive correlation among modality-specific representations across modalities may weaken their complementarity. As a result, an inter-modality orthogonality constraint is further introduced by minimizing the covariance between modality-specific representations from different modalities. The loss is defined as
\begin{equation}
	\mathcal{L}_{ortho2}^M = \sum_{l_1\neq l_2}\left\|\text{Cov}\left(\mathbf{z}^{s_m,l_1,spe}, \mathbf{z}^{s_m,l_2,spe}\right)\right\|_2,
\end{equation}
Minimizing this loss further promotes the specificity of these representations and enhances the complementary nature of heterogeneous sensing information.

The proposed modality-level disentanglement framework is realized by jointly optimizing the three parts of losses, represented as
\begin{equation}
	\mathcal{L}_m = \mathcal{L}_{sim}^M + \mathcal{L}_{ortho1}^M + \mathcal{L}_{ortho2}^M.
\end{equation}
It will enable effective decoupling of shared and specific information across multiple modalities. The disentangled representations could provide a foundation for multi-modal fusion and subsequent domain-level disentanglement.

\subsection{Triple-modal fusion module}
\begin{figure}
	\centering
	\includegraphics[width=0.9\linewidth]{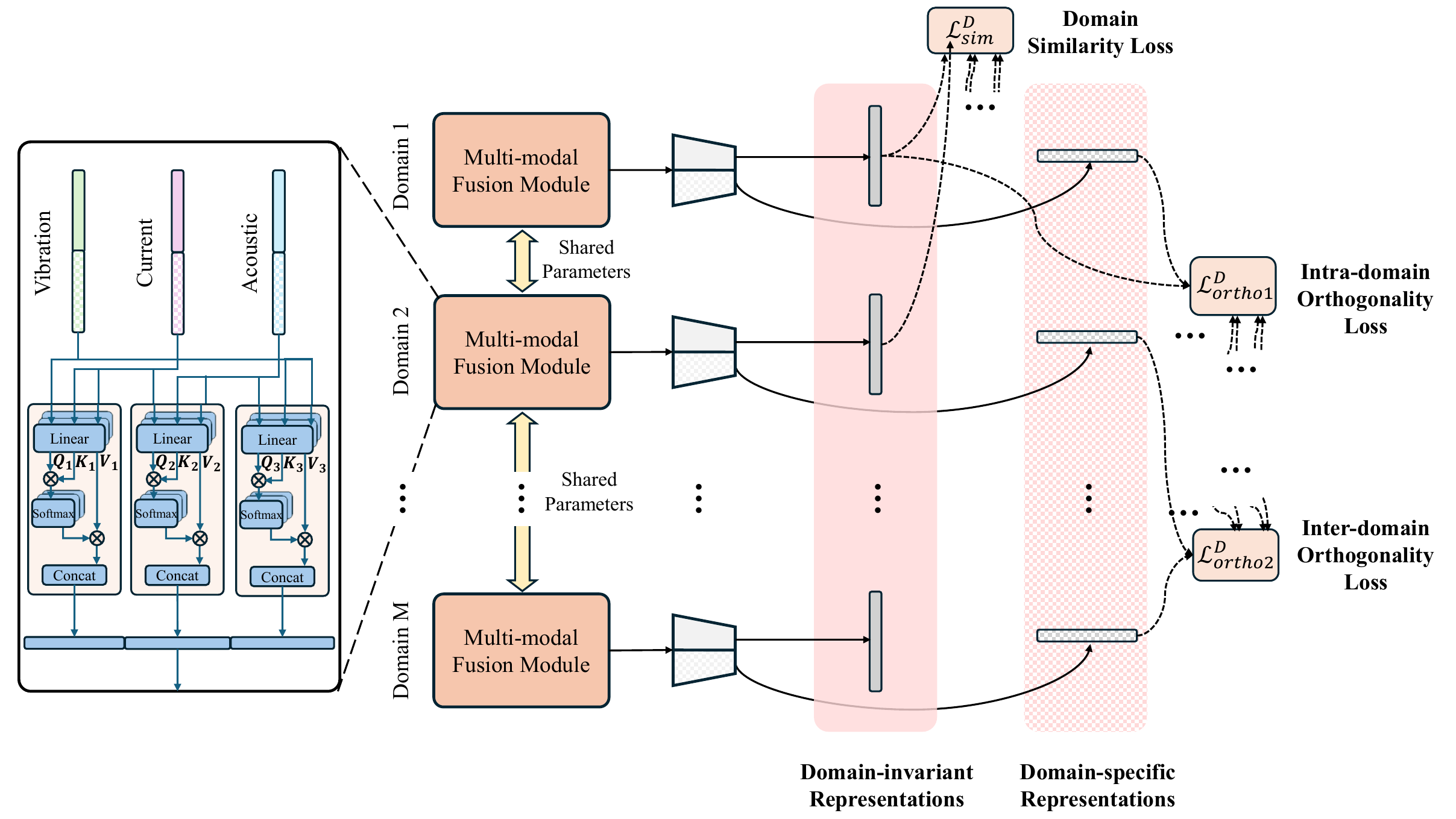}
	\caption{Illustration of triple-modal fusion and domain-level disentanglement.}
	\label{fig:domain-level}
\end{figure}
After modality-level feature disentanglement, the modality-invariant and modality-specific representations from all three modalities are integrated through a multi-modal fusion module based on a multi-head cross-attention mechanism (CAM), as illustrated in Fig.~\ref{fig:domain-level}. Instead of directly performing joint fusion over all three modalities, we follow a pairwise fusion paradigm~\cite{low_rank}, where each pair of heterogeneous modalities is fused independently. This design allows the model to explicitly capture fine-grained inter-modal interactions while avoiding excessive complexity and information interference that may arise from direct triple-modal fusion.

Firstly, for each modality pair $l_p$ and $l_q$, the modality-invariant and modality-specific representations are first concatenated to form the modality-aware embedding
\begin{equation}
	\mathbf{z}^{s_m,l}=\left[\mathbf{z}^{s_m,l,inv}, \mathbf{z}^{s_m,l,spe}\right], l\in\{l_p,l_q\}.
\end{equation}
The query, key, and value matrices in CAM are linearly projected into $H$ attention heads as
\begin{equation}
	\mathbf{Q}_h^{s_m,l_p}=\mathbf{z}^{s_m,l_p}\mathbf{W}_{Q,h}^{(l_p)},
	\mathbf{K}_h^{s_m,l_q}=\mathbf{z}^{s_m,l_q}\mathbf{W}_{K,h}^{(l_q)},
	\mathbf{V}_h^{s_m,l_q}=\mathbf{z}^{s_m,l_q}\mathbf{W}_{V,h}^{(l_q)},
\end{equation}
where $\mathbf{W}_{Q,h}^{(\cdot)}$, $\mathbf{W}_{K,h}^{(\cdot)}$, and $\mathbf{W}_{V,h}^{(\cdot)}$ are learnable projection matrices for the $h$-th head. The fused output is subsequently computed as
\begin{equation}
	\hat{\mathbf{z}}_h^{s_m,(l_p,l_q)}=\text{softmax}\left(\dfrac{\mathbf{Q}_h^{s_m,l_p}\left(\mathbf{K}_h^{s_m,l_q}\right)^T}{\sqrt{d_k}}\right)\mathbf{V}_h^{s_m,l_q},
\end{equation}
where $d_k$ denotes the dimensionality of the key vectors. This operation allows modality $l_p$ to selectively attend to informative components of modality $l_q$, thereby capturing cross-modal dependencies and complementary fault-related patterns. Three modality pairs, i.e., vibration and current, current and acoustic, and acoustic and vibration are fused through this process, respectively, and the three fused embeddings are subsequently concatenated as the comprehensive representations of the input from a single domain, formulated as
\begin{equation}
	\hat{\mathbf{z}}^{s_m} = \left[\text{Concat}\left(\hat{\mathbf{z}}_1^{s_m,(v,c)},\ldots,\hat{\mathbf{z}}_H^{s_m,(v,c)}\right), \text{Concat}\left(\hat{\mathbf{z}}_1^{s_m,(c,a)},\ldots,\hat{\mathbf{z}}_H^{s_m,(c,a)}\right), \text{Concat}\left(\hat{\mathbf{z}}_1^{s_m,(a,v)},\ldots,\hat{\mathbf{z}}_H^{s_m,(a,v)}\right)\right].
\end{equation}

\subsection{Domain-level disentanglement}
Since samples from different source domains originate from distinct working conditions but share the same fault label space, the obtained representations fused multi-modal information contain both domain-invariant information that is shared across different working conditions and domain-specific characteristics that are correlated with particular operating conditions.
Therefore, we propose to extract fault-discriminative features that are insensitive to domain variations while preserving domain-specific information that reflects working condition characteristics from the multi-modal fused representations. To this end, a domain-level disentanglement framework is further introduced to explicitly decouple domain-invariant and domain-specific representations across multiple source domains, as illustrated in Fig.~\ref{fig:domain-level}. Similar to the modality-level design, the fused representation of each sample is projected into two complementary subspaces through two separate embedding networks, yielding a domain-invariant representation and a domain-specific representation, respectively, formulated as
\begin{equation}
	\mathbf{h}_i^{s_m,inv}=F_{inv}\left(\hat{\mathbf{z}}_i^{s_m}\right),
	\mathbf{h}_i^{s_m,spe}=F_{spe}\left(\hat{\mathbf{z}}_i^{s_m}\right),
\end{equation}
where $F_{inv}$ and $F_{spe}$ denote the domain-invariant and domain-specific embedding networks, respectively. Three sets of domain-level disentanglement loss functions are subsequently designed as follows.

\textbf{(1) Domain similarity loss.} The domain-invariant representations from different source domains are expected to follow similar distributions, as they encode fault-related semantics that should generalize across unseen working conditions. To enforce this property, a domain-level MMD loss is introduced to align the invariant feature distributions among all source domains. Formally, the domain similarity loss is defined as
\begin{equation}
	\begin{aligned}
		\mathcal{L}_{sim}^D &= \sum_{m\neq n}\text{MMD}\left(\mathbf{h}_i^{s_m,inv}, \mathbf{h}_i^{s_n,inv}\right)\\
		&=\sum_{m\neq n}\left[\frac{1}{(N_m^s)^2}\sum_{i=1}^{N_m^s}\sum_{j=1}^{N_m^s}k\left(\mathbf{h}_i^{s_m,inv}, \mathbf{h}_j^{s_m,inv}\right) + \frac{1}{(N_n^s)^2}\sum_{i=1}^{N_n^s}\sum_{j=1}^{N_n^s}k\left(\mathbf{h}_i^{s_n,inv}, \mathbf{h}_j^{s_n,inv}\right)\right.\\
		&\phantom{=\;\;}\left.- \frac{2}{N_m^sN_n^s}\sum_{i=1}^{N_m^s}\sum_{j=1}^{N_n^s}k\left(\mathbf{h}_i^{s_m,inv}, \mathbf{z}_j^{s_n,inv}\right)\right].
	\end{aligned}
\end{equation}
By minimizing this loss, the model is encouraged to learn domain-agnostic representations that are robust to variations of working condition.

\textbf{(2) Intra-domain orthogonality loss.} While invariant representations capture shared fault semantics, domain-specific representations are intended to model working-condition-related characteristics. To prevent redundancy between these two subspaces, an orthogonality constraint is imposed within each domain by minimizing the statistical dependence between domain-invariant and domain-specific representations. Specifically, a covariance-based orthogonality loss is formulated as
\begin{equation}
	\mathcal{L}_{ortho1}^D = \sum_{m=1}^{M}\left\|\text{Cov}\left(\mathbf{h}_i^{s_m,inv}, \mathbf{h}_i^{s_m,spe}\right)\right\|_2.
\end{equation}
This loss facilitates a clear separation between the domain-invariant subspace and the domain-specific subspace to prevent information leakage.

\textbf{(3) Inter-domain orthogonality loss.} In addition, domain-specific representations from different source domains should remain distinctive, as each domain corresponds to a unique working condition. There should be limited correlation among domain-specific features across domains. As a result, an inter-domain orthogonality constraint is introduced to minimize the covariance between domain-specific representations from different source domains, defined as
\begin{equation}
	\mathcal{L}_{ortho2}^D = \sum_{m\neq n}\left\|\text{Cov}\left(\mathbf{h}_i^{s_m,spe}, \mathbf{h}_i^{s_n,spe}\right)\right\|_2.
\end{equation}

The proposed domain-level disentanglement is achieved by jointly optimizing the domain similarity loss, the intra-domain orthogonality loss, and the inter-domain orthogonality loss, formulated as follows.
\begin{equation}
	\mathcal{L}_d = \mathcal{L}_{sim}^D + \mathcal{L}_{ortho1}^D + \mathcal{L}_{orth2}^D.
\end{equation}
This optimization enables the separation of shared and specific features across source working conditions, thereby enhancing generalization towards unseen working conditions. Combined with the modality-level disentanglement and multi-modal fusion modules, the ultimate representations are utilized for fault diagnosis.

\subsection{Overall objective function}
The learned domain-invariant and domain-specific representations are concatenated as input of a fault classifier to obtain the predicted fault probability $\hat{y}_i^{s_m}$. To ensure discriminative capability for fault diagnosis, a cross-entropy loss is employed on the source domain samples. The classification loss is formulated as
\begin{equation}
	\mathcal{L}_{cls}=\frac{1}{M}\sum_{m=1}^{M}\frac{1}{N_m^s}\text{CE}\left(\hat{y}_i^{s_m}, y_i^{s_m}\right),
\end{equation}
where $\text{CE}$ denotes the cross-entropy loss.

The proposed framework is trained in an end-to-end manner by jointly optimizing the fault classification objective and the dual-level disentanglement loss. The overall loss function is defined as a weighted combination of these three parts, represented as
\begin{equation}
	\mathcal{L} = \mathcal{L}_{cls} + \lambda_m\mathcal{L}_m + \lambda_d\mathcal{L}_d,
\end{equation}
where $\lambda_m$ and $\lambda_d$ are non-negative trade-off parameters for the modality-level disentanglement and the domain-level disentanglement, respectively. After training on multiple source domains, the well-trained model is directly used in the target domain under unseen working conditions without any samples involved in the training process.

\section{Experimental verification}\label{sec_experiment}
\subsection{Dataset construction}
\begin{figure}
	\centering
	\includegraphics[width=0.75\linewidth]{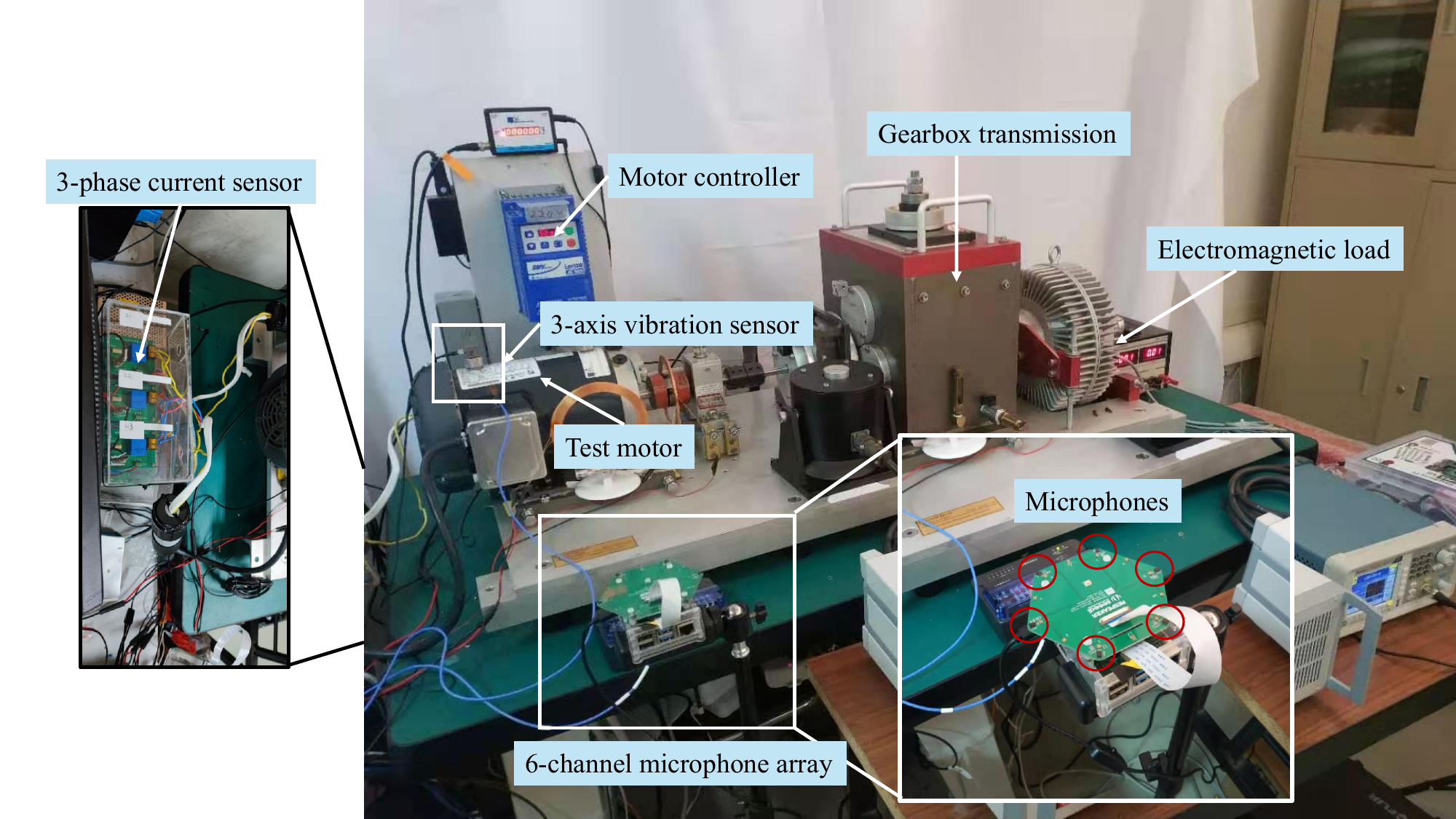}
	\caption{Induction motor fault experiment platform.}
	\label{fig:platform}
\end{figure}

To validate the proposed multi-modal DG method for fault diagnosis under unseen conditions, an induction motor fault dataset was constructed using a Drivetrain Dynamics Simulator (DDS) platform, as shown in Fig.~\ref{fig:platform}. The test motor drives a gearbox transmission system that is coupled to an electromagnetic load, while the rotating speed of the motor is precisely regulated by a motor controller. Multi-modal sensing signals are synchronously acquired during operation. Specifically, a triaxial vibration sensor is mounted on the motor housing to capture motor vibration signals along three orthogonal directions. Three-phase stator current signals are collected using three current sensors. In addition, a microphone array consisting of six microphones is deployed in proximity to the test motor to record acoustic signals. The vibration and current signals are sampled at a frequency of 5,120~Hz, and the microphones have a sampling rate of 44,100~Hz.

\begin{table}[width=.6\linewidth,cols=3]
	\caption{Categories of motor health states.}\label{tab:faults}
	\setlength{\tabcolsep}{14pt}
	\begin{tabular*}{\tblwidth}{ccc}
		\toprule
		Class label & Fault type & Abbreviation\\
		\midrule
		1 & Normal & N\\
		2 & Broken rotor bar & BRB\\
		3 & Stator winding fault & SWF\\
		4 & Parallel misaligned rotor & PMR\\
		5 & Bearing fault & BF\\
		6 & Rotor bow & RB\\
		7 & Angular misaligned rotor & AMR\\
		8 & Rotor unbalance & RU\\
		\bottomrule
	\end{tabular*}
\end{table}
Eight test motors with different health states were employed in the experiments, including one healthy motor (N) and seven faulty motors, as summarized in Table~\ref{tab:faults}. The broken rotor bar (BRB) motor contains three fractured bars in the rotor cage. The stator winding fault (SWF) motor exhibits inter-turn short-circuit faults in the stator windings. The parallel misaligned rotor (PMR) fault corresponds to a radial displacement of the rotor relative to the motor centerline while maintaining parallel alignment with the shaft axis. In contrast, the angular misaligned rotor (AMR) fault is characterized by a radial displacement occurring at only one end of the rotor. For the bearing fault (BF) motor, two bearings of the motor have an inner race defect and an outer race defect, respectively. The rotor bow (RB) fault indicates that the rotor shaft is permanently bent, while the rotor unbalance (RU) fault is introduced by attaching unbalanced masses to the rotor to generate asymmetric centrifugal forces.

\begin{figure}
	\centering
	\subfigure[]{
		\begin{minipage}{0.30\linewidth}
			\centering
			\includegraphics[width=\linewidth]{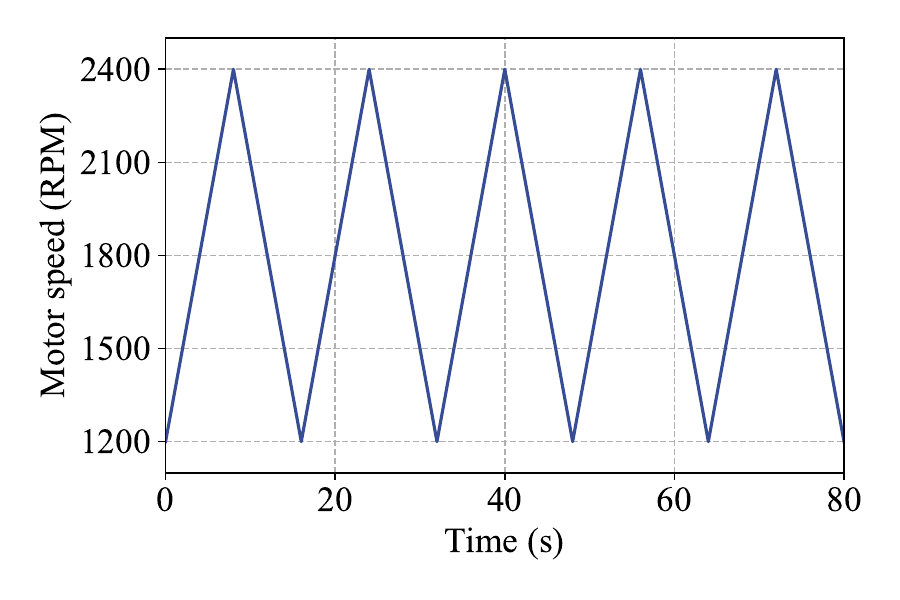}
		\end{minipage}
	}
	\subfigure[]{
		\begin{minipage}{0.30\linewidth}
			\centering
			\includegraphics[width=\linewidth]{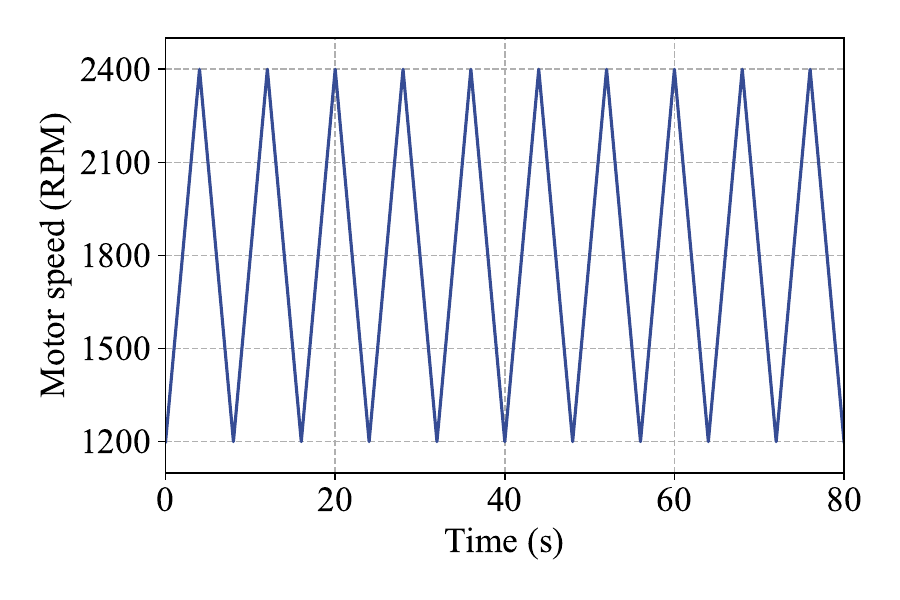}
		\end{minipage}
	}
	\subfigure[]{
		\begin{minipage}{0.30\linewidth}
			\centering
			\includegraphics[width=\linewidth]{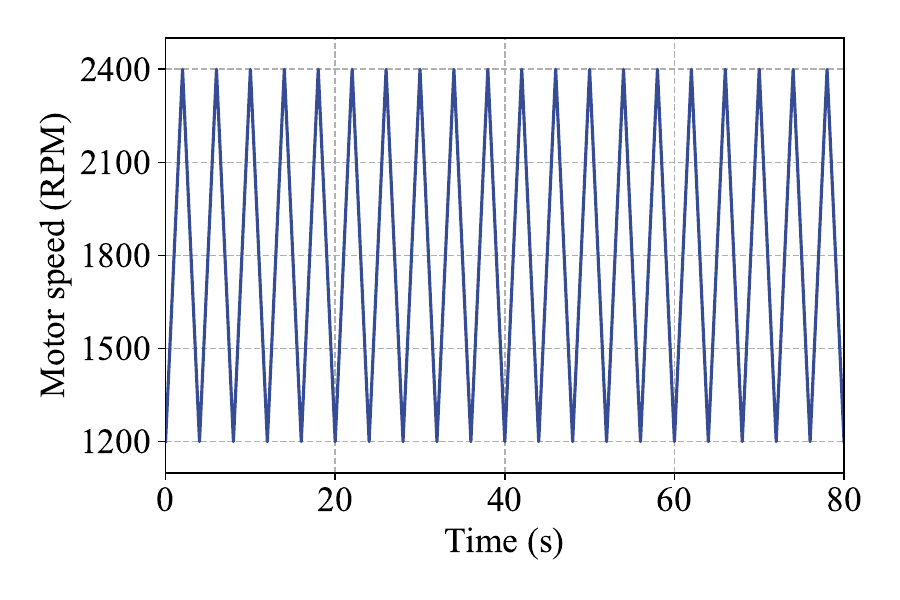}
		\end{minipage}
	}
	\subfigure[]{
		\begin{minipage}{0.30\linewidth}
			\centering
			\includegraphics[width=\linewidth]{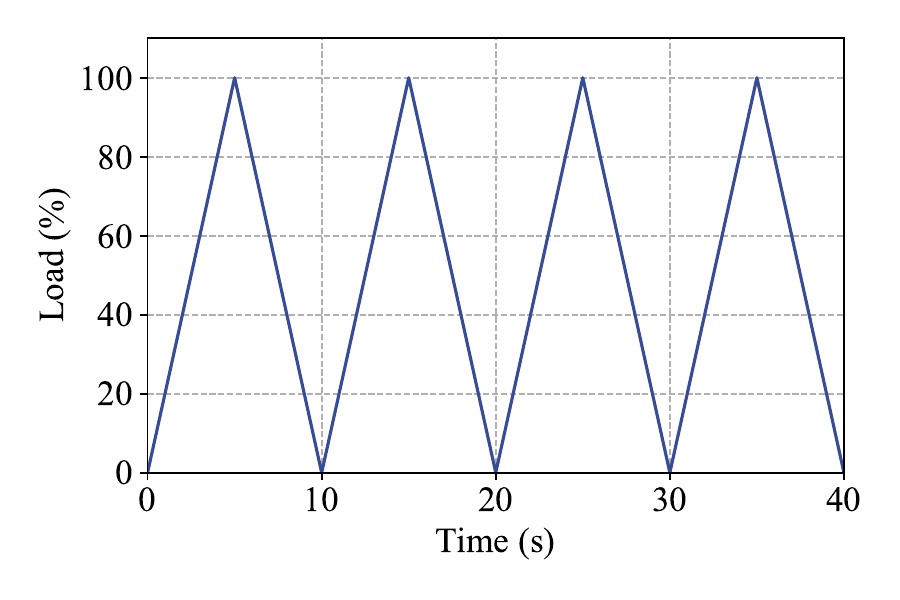}
		\end{minipage}
	}
	\subfigure[]{
		\begin{minipage}{0.30\linewidth}
			\centering
			\includegraphics[width=\linewidth]{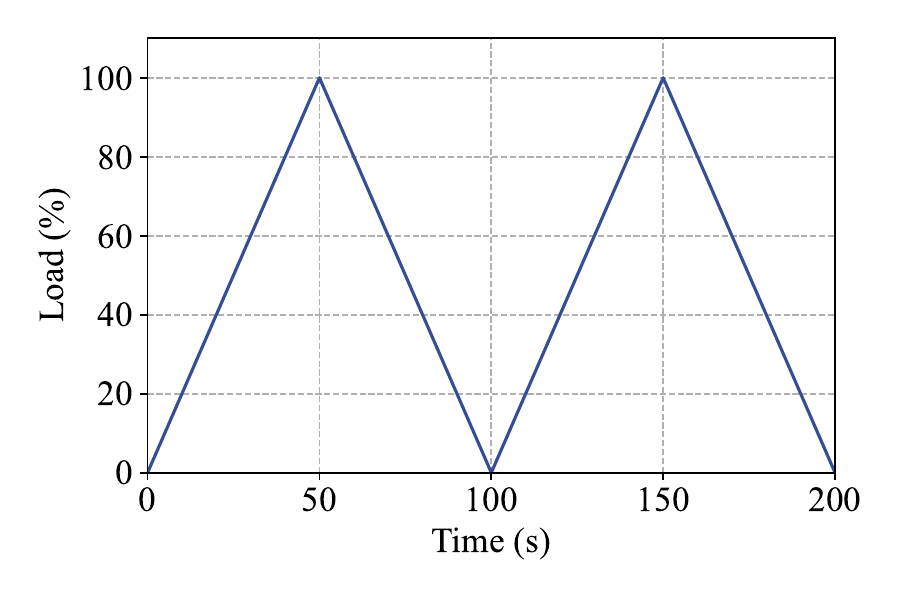}
		\end{minipage}
	}
	\caption{Rotating speed curves of working condition (a) C5, (b) C6, (c) C7, and load curves of working condition (d) C8, (e) C9.}
	\label{fig:motor_condition}
\end{figure}

All eight motors were tested under nine distinct working conditions, including four constant-speed conditions and five time-varying conditions. Detailed descriptions of these operating conditions are provided in Table~\ref{tab:conditions}. The first four conditions (C1-C4) correspond to constant but different rotating speeds, while sharing the same load level, which is set to 100\% of the platform's maximum load of 72~N·m. Conditions C5, C6, and C7 involve cyclic speed variations between 1200~RPM and 2400~RPM with different acceleration and deceleration rates, as illustrated in Fig.~\ref{fig:motor_condition}(a)-(c), while the load remains constant. In contrast, conditions C8 and C9 are characterized by a constant rotating speed of 1800~RPM, whereas the load follows a cyclic increasing-and-decreasing pattern between 0\% and 100\%, as shown in Fig.~\ref{fig:motor_condition}(d) and (e). Notably, data acquisition was initiated 10~s after motor startup to ensure stable operating conditions.
\begin{table}[width=.8\linewidth,cols=5]
	\caption{Details of working conditions in the experiments.}\label{tab:conditions}
	\setlength{\tabcolsep}{12pt}
	\begin{tabular*}{\tblwidth}{ccccc}
		\toprule
		Condition & Speed (RPM) & \makecell{Change rate of \\speed (RPM/s)} & Load percentage (\%) & \makecell{Change rate of \\load (\%/s)}\\
		\midrule
		C1 & 1200 & 0 & 100 & 0\\
		C2 & 1800 & 0 & 100 & 0\\
		C3 & 2400 & 0 & 100 & 0\\
		C4 & 2700 & 0 & 100 & 0\\
		C5 & 1200-2400 & 150 & 100 & 0\\
		C6 & 1200-2400 & 300 & 100 & 0\\
		C7 & 1200-2400 & 600 & 100 & 0\\
		C8 & 1800 & 0 & 0-100 & 20\\
		C9 & 1800 & 0 & 0-100 & 2\\
		\bottomrule
	\end{tabular*}
\end{table}

\subsection{Implementation details}
For each working condition and each motor, the collected multi-modal signals were segmented into fixed-length samples of 0.2~s using a non-overlapping sliding window. The first 100~s of data under stable operating conditions were selected for analysis, resulting in 500 samples for each fault category under each working condition. Based on this dataset, two schemes of cross-condition fault diagnosis tasks were designed, as summarized in Table~\ref{tab:tasks}. The first type (T1-T4) corresponds to scenarios in which the model is trained using data from multiple constant working conditions and evaluated on a different constant condition. The second type (T5-T9) represents a more challenging and practically relevant setting, where the model is trained on data from multiple constant working conditions and tested on time-varying operating conditions.
\begin{table}[width=.6\linewidth,cols=3]
	\caption{Details of cross-condition tasks.}\label{tab:tasks}
	\setlength{\tabcolsep}{12pt}
	\begin{tabular*}{\tblwidth}{ccc}
		\toprule
		Task & Source working condition & Target working condition\\
		\midrule
		T1 & C2, C3, C4 & C1\\
		T2 & C1, C3, C4 & C2\\
		T3 & C1, C2, C4 & C3\\
		T4 & C1, C2, C3 & C4\\
		T5 & C1, C2, C3 & C5\\
		T6 & C1, C2, C3 & C6\\
		T7 & C1, C2, C3 & C7\\
		T8 & C1, C2, C3 & C8\\
		T9 & C1, C2, C3 & C9\\
		\bottomrule
	\end{tabular*}
\end{table}

In this work, vibration signals were transformed into time-frequency representations using STFT with a Hann window, a Fourier transform length of 127, and a hop length of 33, producing time-frequency maps of size $64 \times 32 \times 3$. Acoustic signals were converted into Mel-spectrograms with a Fourier transform length of 512, a hop length of 138, and 64 Mel frequency bins, resulting in representations of size $64 \times 64 \times 6$. The current signals were kept in their original form with a size of $1024 \times 3$. For feature extraction, all signal encoders adopt a ResNet architecture consisting of a convolutional preprocessing layer followed by four residual blocks. Specifically, 2D ResNets are employed for vibration and acoustic modalities, while a 1D ResNet is used for the current modality. Each encoder consists of an initial convolutional layer followed by four residual stages with gradually increasing channel dimensions (from 16 to 256). Each stage contains one residual block, composing of two convolutional layers with skip connections. For the 1D encoder, all convolutional layers are implemented in 1D form, and the first layer adopts a wide kernel with a size of 64 to capture long-range temporal dependencies in current signals. All the encoders employ a global average pooling (GAP) layer to obtain the representation embedding with a dimension of 256. Both the modality-level and domain-level embedding networks are implemented as single-layer fully connected networks, and the fault classifier also adopts a single-layer fully connected structure. The dimensionalities of the modality-invariant, modality-specific, domain-invariant, and domain-specific representations are all set to 128. In the triple-modal fusion module, 8 heads with dimensionality of 32 are used for CAM. The model is trained using the Adam optimizer with a learning rate of 0.001 and a batch size of 256 for each source domain. All experiments are conducted on an NVIDIA GeForce RTX 4090 GPU and repeated five times to mitigate the effects of randomness.

Nine state-of-the-art methods are introduced for comparison, including six representative DG approaches and three multi-modal fusion methods for fault diagnosis :
\begin{enumerate}[leftmargin=2\parindent, noitemsep]
	\item[1)] \textit{MMD}: A basic DG method which aligns feature distributions among multiple source domains by minimizing the summation of MMD across all domain pairs.
	\item[2)] \textit{DANN}~\cite{DANN}: A domain adversarial neural network which uses adversarial learning for multi-source domain alignment.
	\item[3)] \textit{MLDG}~\cite{MLDG}: A meta-learning-based DG method that learns domain-generalizable representations by simulating domain shifts during training.
	\item[4)] \textit{CDDG}~\cite{CDDG}: A causal disentanglement DG method that separates causal factors from non-causal factors to enhance generalization.
	\item[5)] \textit{DTDG}~\cite{adversarial2}: A domain transferability-based DG method assigning weights to source domains in adversarial learning.
	\item[6)] \textit{CCN}~\cite{CCN}: A causal consistency network-based DG method that introduces a causal consistency loss to model the causality of faults.
	\item[7)] \textit{FCFN}~\cite{feature_fusion}: A multi-sensor fusion network with a feature convergence layer to integrate heterogeneous sensor information.
	\item[8)] \textit{MFHR}~\cite{MFHR}: A multi-sensor fusion method for high-reliability fault diagnosis with a decision-level fusion.
	\item[9)] \textit{MMCDF}~\cite{MMDA}: A multi-modal cross-domain fusion network that integrates DA with multi-modal feature fusion.
\end{enumerate}
For fair comparison, all methods are implemented using the same backbone feature extractors as the proposed method. Notably, MMD and DANN were originally developed under the DA setting. To adapt them to the DG setting, the original source-target distribution alignment is replaced by pairwise alignment among all source domains for MMD~\cite{MMD}, and the domain discriminator is extended to a multi-class source-domain discriminator for DANN~\cite{DANN}. Similarly, MMCDF is adapted by replacing its source-target MMD alignment with the summation of pairwise MMD losses among all source domains. In DG methods designed for single-modal data, the features from multiple modalities are concatenated as the comprehensive representations.

\subsection{Comparison results}
\begin{figure}
	\centering
	\subfigure[]{
		\begin{minipage}{0.48\linewidth}
			\centering
			\includegraphics[width=\linewidth]{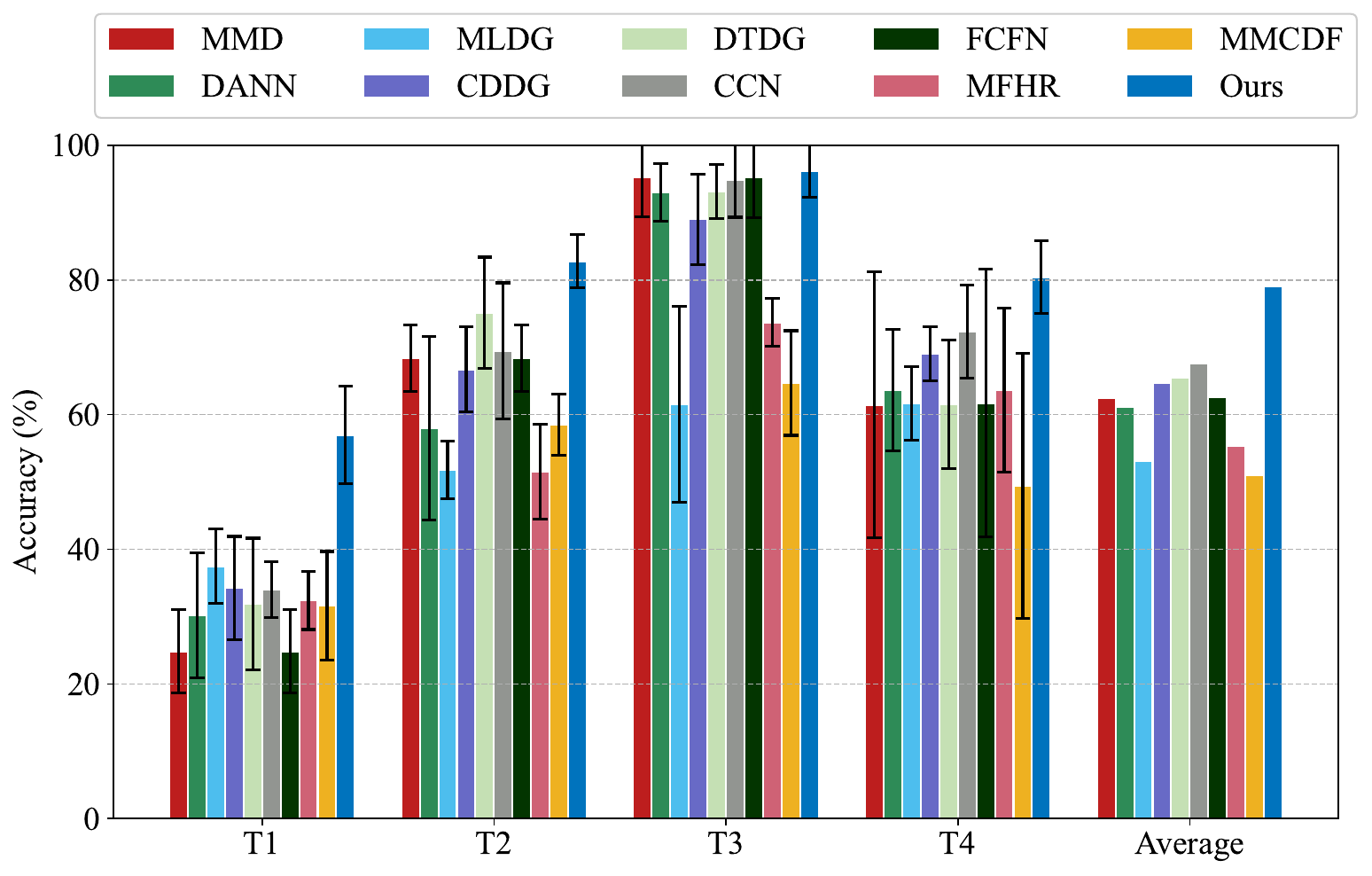}
		\end{minipage}
	}
	\subfigure[]{
		\begin{minipage}{0.48\linewidth}
			\centering
			\includegraphics[width=\linewidth]{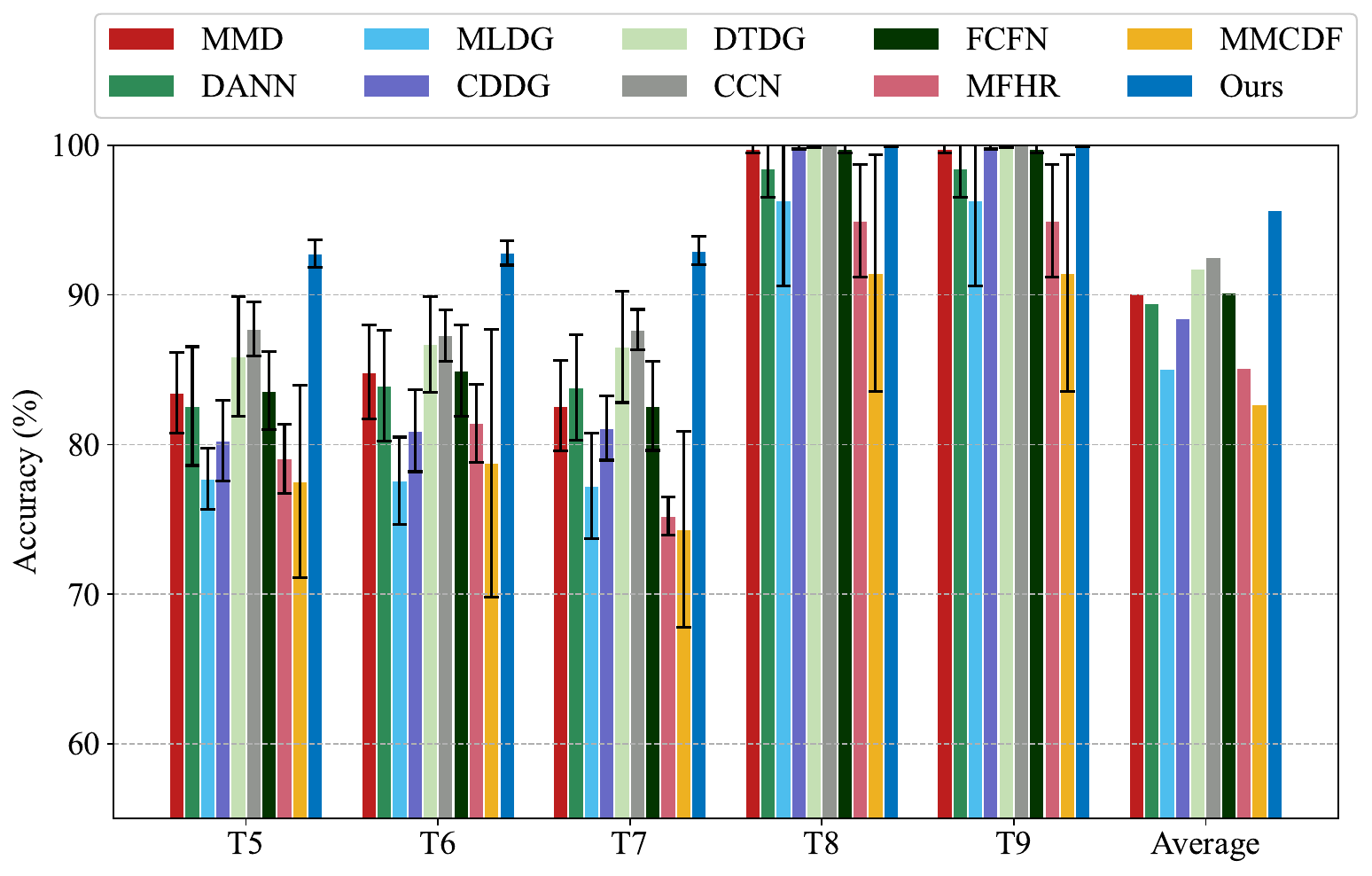}
		\end{minipage}
	}
	\caption{Fault diagnosis accuracy of the comparative methods. (a) Task scheme 1. (b) Task scheme 2.}
	\label{fig:results}
\end{figure}
The diagnosis results of the proposed method (Ours) and the comparison methods under the first scheme of cross-condition tasks are illustrated in Fig.~\ref{fig:results}(a), and the average accuracy is also calculated. Overall, the proposed method consistently outperforms all comparison methods across T1-T4 as well as in terms of average accuracy, demonstrating its superior generalization capability under unseen working conditions. It can be observed that the accuracies achieved on Tasks T2 and T3 are higher than those on Tasks T1 and T4. It suggests that the rotating speeds of the target domains in T2 and T3 are closer to, or fall within, the range of the source domains, leading to reduced domain discrepancy. Specifically, traditional DG methods, such as MMD and DANN, exhibit relatively poor performance, particularly in T1, where the target condition shows a large distribution discrepancy from the source domains, as the rotating speed of 1200 RPM is much lower than that in the source domains. This indicates that enforcing global domain invariance alone is insufficient to handle complex condition shifts in fault diagnosis scenarios. Although methods such as CDDG, DTDG, and CCN improve performance by incorporating causal factors or weighting strategies, their gains remain limited, suggesting that condition-related specific information is still inadequately modeled. Furthermore, these methods are designed for single-modal data and cannot exploit complementary information across multiple sensing modalities, which limits their generalization capability. Multi-modal fusion methods, such as FCFN and MFHR, outperform several DG baselines, demonstrating the benefit of integrating multiple sensors. Nevertheless, these approaches mainly focus on information fusion under constant conditions and do not explicitly address domain generalization. MMCDF achieves better performance by combining multi-modal fusion with DA, while it still achieves inferior performance compared with the proposed method.

Fig.~\ref{fig:results}(b) reports the fault diagnosis results on the second task scheme, where the models are trained on multiple constant working conditions and tested on unseen time-varying conditions. All the methods have relatively higher accuracies than those obtained in most tasks in the first scheme, especially in tasks T8 and T9. This suggests that load variations may induce relatively smaller domain shifts compared with speed variations, and that exposure to constant-speed source domains within the target speed range can facilitate generalization to time-varying operating conditions. Traditional DG methods show limited robustness under time-varying conditions, as their domain alignment strategies mainly rely on global distribution matching and struggle to capture dynamically changing characteristics. Advanced DG approaches, particularly for DTDG and CCN, achieve relatively better performance by incorporating domain weighting or causal modeling, yet their generalization ability remains constrained due to the simple multi-modal fusion strategy. Multi-modal fusion methods show unstable diagnosis performance, demonstrating their limitations to tackle domain shifts under unseen working conditions. In contrast, the proposed method consistently achieves the highest accuracy across almost all time-varying tasks and yields the best average performance. Furthermore, the proposed method demonstrates lower standard deviations compared with other methods, indicating its high statistical reliability. This superiority can be attributed to the synergistic effects of cross-domain mixed fusion and dual disentanglement, which effectively enhance robustness to dynamic domain shifts while preserving both modality-specific and domain-related discriminative information.

\subsection{Ablation study}
To evaluate the contribution of each component in the proposed framework, comprehensive ablation studies were conducted by selectively removing individual modules while keeping the remaining architecture consistent with the original model. Specifically, \textit{w/o modality-dis}, \textit{w/o domain-dis}, and \textit{w/o dis} denote the variants in which the modality-level disentanglement, domain-level disentanglement, and both dual-level disentanglement are removed, respectively. These variants are implemented by setting the corresponding trade-off coefficients in the overall loss function to zero. The variant \textit{w/o mix} indicates that the proposed cross-domain mixed fusion mechanism is disabled. To further investigate the effectiveness of the proposed triple-modal fusion module, several alternative fusion strategies are considered. The variant \textit{concat} removes the proposed fusion module and directly concatenates the representations from the three modalities. In \textit{concat\_emb}, an additional embedding layer consisting of a single fully-connected layer is applied to the concatenated features. Similarly, element-wise feature addition across modalities is introduced as another fusion strategy, denoted as \textit{add}, while \textit{add\_emb} further applies a single fully-connected embedding layer after the addition operation. Finally, \textit{Baseline} represents the simplest configuration, in which all the aforementioned components are removed, and multi-modal features are fused solely by direct concatenation. The results of these ablation methods are presented in Table~\ref{tab:ablation1}.
\begin{table}[width=\linewidth,cols=11]
	\caption{Accuracy of ablation study on different tasks (\%).}\label{tab:ablation1}
	\setlength{\tabcolsep}{8pt}
	\renewcommand{\arraystretch}{1.2}
	\begin{tabular*}{\tblwidth}{ccccccccccc}
		\toprule
		Method & T1 & T2 & T3 & T4 & T5 & T6 & T7 & T8 & T9 & Average\\
		\midrule
		Baseline & \makecell{$36.98$\\{\footnotesize $\pm 15.66$}} & \makecell{$68.10$\\{\footnotesize $\pm 6.17$}} & \makecell{$83.00$\\{\footnotesize $\pm5.52$}} & \makecell{$51.16$\\{\footnotesize $\pm6.18$}} & \makecell{$78.52$\\{\footnotesize $\pm3.15$}} & \makecell{$80.04$\\{\footnotesize $\pm3.95$}} & \makecell{$79.05$\\{\footnotesize $\pm3.25$}} & \makecell{$98.53$\\{\footnotesize $\pm1.49$}} & \makecell{$98.53$\\{\footnotesize $\pm1.49$}} & 74.88\\
		w/o dis & \makecell{51.18\\{\footnotesize $\pm9.53$}} & \makecell{80.65\\{\footnotesize $\pm7.88$}} & \makecell{92.73\\{\footnotesize $\pm6.29$}} & \makecell{72.86\\{\footnotesize $\pm1.87$}} & \makecell{90.84\\{\footnotesize $\pm2.19$}} & \makecell{91.08\\{\footnotesize $\pm2.04$}} & \makecell{91.12\\{\footnotesize $\pm1.65$}} & \makecell{99.97\\{\footnotesize $\pm0.04$}} & \makecell{99.97\\{\footnotesize $\pm0.04$}} & 85.60\\
		w/o modality-dis & \makecell{54.20\\{\footnotesize $\pm3.36$}} & \makecell{83.00\\{\footnotesize $\pm4.46$}} & \makecell{94.28\\{\footnotesize $\pm3.25$}} & \makecell{79.21\\{\footnotesize $\pm4.90$}} & \makecell{93.18\\{\footnotesize $\pm1.36$}} & \makecell{93.23\\{\footnotesize $\pm1.32$}} & \makecell{93.65\\{\footnotesize $\pm1.35$}} & \makecell{99.95\\{\footnotesize $\pm0.03$}} & \makecell{99.95\\{\footnotesize $\pm0.03$}} & 87.85\\
		w/o domain-dis & \makecell{53.70\\{\footnotesize $\pm8.52$}} & \makecell{80.68\\{\footnotesize $\pm2.24$}} & \makecell{92.02\\{\footnotesize $\pm7.52$}} & \makecell{76.35\\{\footnotesize $\pm5.72$}} & \makecell{91.31\\{\footnotesize $\pm1.67$}} & \makecell{91.28\\{\footnotesize $\pm1.15$}} & \makecell{91.60\\{\footnotesize $\pm1.07$}} & \makecell{99.87\\{\footnotesize $\pm0.21$}} & \makecell{99.87\\{\footnotesize $\pm0.21$}} & 86.30\\
		w/o mix & \makecell{45.40\\{\footnotesize $\pm10.86$}} & \makecell{75.99\\{\footnotesize $\pm7.66$}} & \makecell{93.70\\{\footnotesize $\pm4.81$}} & \makecell{69.99\\{\footnotesize $\pm7.73$}} & \makecell{85.01\\{\footnotesize $\pm2.86$}} & \makecell{85.92\\{\footnotesize $\pm2.60$}} & \makecell{85.43\\{\footnotesize $\pm2.60$}} & \makecell{99.78\\{\footnotesize $\pm0.19$}} & \makecell{99.78\\{\footnotesize $\pm0.19$}} & 82.33\\
		concat & \makecell{52.54\\{\footnotesize $\pm12.36$}} & \makecell{68.59\\{\footnotesize $\pm4.32$}} & \makecell{86.94\\{\footnotesize $\pm6.77$}} & \makecell{79.51\\{\footnotesize $\pm4.51$}} & \makecell{90.99\\{\footnotesize $\pm1.33$}} & \makecell{91.72\\{\footnotesize $\pm1.77$}} & \makecell{92.35\\{\footnotesize $\pm1.78$}} & \makecell{99.93\\{\footnotesize $\pm0.07$}} & \makecell{99.93\\{\footnotesize $\pm0.07$}} & 84.72\\
		concat\_emb & \makecell{50.16\\{\footnotesize $\pm9.51$}} & \makecell{75.82\\{\footnotesize $\pm13.81$}} & \makecell{91.74\\{\footnotesize $\pm9.67$}} & \makecell{78.15\\{\footnotesize $\pm3.41$}} & \makecell{91.40\\{\footnotesize $\pm1.12$}} & \makecell{91.76\\{\footnotesize $\pm0.93$}} & \makecell{91.69\\{\footnotesize $\pm1.48$}} & \makecell{99.98\\{\footnotesize $\pm0.02$}} & \makecell{99.98\\{\footnotesize $\pm0.02$}} & 85.63\\
		add & \makecell{43.10\\{\footnotesize $\pm10.67$}} & \makecell{75.89\\{\footnotesize $\pm4.79$}} & \makecell{93.68\\{\footnotesize $\pm4.96$}} & \makecell{78.65\\{\footnotesize $\pm4.80$}} & \makecell{90.19\\{\footnotesize $\pm2.13$}} & \makecell{90.31\\{\footnotesize $\pm1.96$}} & \makecell{91.01\\{\footnotesize $\pm1.62$}} & \makecell{100.00\\{\footnotesize $\pm0.00$}} & \makecell{100.00\\{\footnotesize $\pm0.00$}} & 84.76\\
		add\_emb & \makecell{42.92\\{\footnotesize $\pm14.34$}} & \makecell{73.64\\{\footnotesize $\pm8.50$}} & \makecell{95.02\\{\footnotesize $\pm3.16$}} & \makecell{78.11\\{\footnotesize $\pm4.41$}} & \makecell{91.13\\{\footnotesize $\pm1.34$}} & \makecell{91.39\\{\footnotesize $\pm1.63$}} & \makecell{91.80\\{\footnotesize $\pm1.67$}} & \makecell{100.00\\{\footnotesize $\pm0.01$}} & \makecell{100.00\\{\footnotesize $\pm0.01$}} & 84.89\\
		Ours & \makecell{56.98\\{\footnotesize $\pm7.23$}} & \makecell{82.79\\{\footnotesize $\pm3.96$}} & \makecell{96.24\\{\footnotesize $\pm3.91$}} & \makecell{80.42\\{\footnotesize $\pm5.42$}} & \makecell{92.75\\{\footnotesize $\pm0.93$}} & \makecell{92.82\\{\footnotesize $\pm0.82$}} & \makecell{92.98\\{\footnotesize $\pm0.95$}} & \makecell{99.97\\{\footnotesize $\pm0.04$}} & \makecell{99.97\\{\footnotesize $\pm0.04$}} & 88.32\\
		\bottomrule
	\end{tabular*}
\end{table}

Overall, removing any key module leads to performance degradation compared to the complete model in terms of the average accuracy, demonstrating that all components contribute positively to fault diagnosis generalization. First, the \textit{Baseline} model achieves the lowest average accuracy, indicating that simple multi-modal feature concatenation without mixed fusion or disentanglement is insufficient for handling cross-condition generalization. Introducing disentanglement mechanisms significantly improves performance. Compared to Baseline, the variants \textit{w/o modality-dis} and \textit{w/o domain-dis} both yield notable accuracy gains, suggesting that disentangling modality-related and domain-related factors is crucial for learning transferable representations. When both disentanglement mechanisms are removed (\textit{w/o dis}), the performance further drops, highlighting the complementary nature of dual-level disentanglement. Relatively, domain-level disentanglement contributes more in the proposed framework, as dropping this module results in a higher accuracy decline, indicating the significance of domain-invariant and domain-specific feature learning for cross-domain diagnosis. Second, disabling the cross-domain mixed fusion (\textit{w/o mix}) leads to a significant decrease in average accuracy, which verifies the effectiveness of modality augmentation across source domains in mitigating domain bias. Third, replacing the proposed triple-modal fusion module with simple fusion strategies consistently results in inferior performance. Although adding an embedding layer slightly improves the results compared with direct fusion, these methods still lag behind the proposed cross-attention-based fusion, demonstrating that deep and adaptive modeling of inter-modal correlations is essential for effective multi-modal collaboration. Finally, the proposed method achieves the highest accuracy across most tasks and the best average performance, confirming that the joint design of cross-domain mixed fusion, dual-level disentanglement, and triple-modal fusion is critical for achieving robust fault diagnosis under unseen working conditions.

In addition, experiments were conducted using single-modality information to further verify the advantage of multi-modal data fusion in cross-domain motor fault diagnosis. Specifically, a single encoder was employed to extract features from one modality at a time, while the domain-level disentanglement mechanism was retained for fair comparison. The fault diagnosis results obtained using each individual modality are summarized in Table~\ref{tab:ablation2}. It can be observed that models trained on a single modality exhibit substantially lower accuracy than the multi-modal model across almost all tasks, indicating that relying on a single sensing source is insufficient for robust cross-domain fault diagnosis. Among the single-modality approaches, vibration signals achieve the best overall performance, which can be attributed to their rich fault-related mechanical information. However, their performance still degrades under certain working conditions, especially when the domain shift becomes significant. Current signals and acoustic signals individually show much lower accuracy, reflecting their limited discriminative capability when used in isolation and their higher sensitivity to variations of operating conditions. The proposed multi-modal method achieves the highest accuracy across all tasks, demonstrating that fusing heterogeneous signals enables the model to exploit complementary fault characteristics and improve generalization ability.
\begin{table}[width=\linewidth,cols=11]
	\caption{Accuracy of single-modal models on different tasks (\%).}\label{tab:ablation2}
	\setlength{\tabcolsep}{8pt}
	\renewcommand{\arraystretch}{1.2}
	\begin{tabular*}{\tblwidth}{ccccccccccc}
		\toprule
		Modality & T1 & T2 & T3 & T4 & T5 & T6 & T7 & T8 & T9 & Average\\
		\midrule
		Vibration & \makecell{40.48\\{\footnotesize $\pm7.78$}} & \makecell{72.75\\{\footnotesize $\pm9.56$}} & \makecell{83.12\\{\footnotesize $\pm9.45$}} & \makecell{73.02\\{\footnotesize $\pm3.30$}} & \makecell{75.50\\{\footnotesize $\pm3.51$}} & \makecell{76.55\\{\footnotesize $\pm3.57$}} & \makecell{77.05\\{\footnotesize $\pm2.90$}} & \makecell{99.66\\{\footnotesize $\pm0.40$}} & \makecell{99.66\\{\footnotesize $\pm0.40$}} & 77.53\\
		Current & \makecell{17.32\\{\footnotesize $\pm4.50$}} & \makecell{23.58\\{\footnotesize $\pm10.66$}} & \makecell{22.76\\{\footnotesize $\pm5.43$}} & \makecell{21.62\\{\footnotesize $\pm4.40$}} & \makecell{36.19\\{\footnotesize $\pm3.21$}} & \makecell{36.45\\{\footnotesize $\pm3.22$}} & \makecell{35.12\\{\footnotesize $\pm2.66$}} & \makecell{35.52\\{\footnotesize $\pm6.82$}} & \makecell{35.52\\{\footnotesize $\pm6.82$}} & 29.34\\
		Acoustic & \makecell{20.42\\{\footnotesize $\pm4.26$}} & \makecell{33.86\\{\footnotesize $\pm6.86$}} & \makecell{60.90\\{\footnotesize $\pm14.67$}} & \makecell{30.08\\{\footnotesize $\pm8.05$}} & \makecell{50.16\\{\footnotesize $\pm5.94$}} & \makecell{45.79\\{\footnotesize $\pm5.93$}} & \makecell{44.80\\{\footnotesize $\pm5.77$}} & \makecell{49.13\\{\footnotesize $\pm8.40$}} & \makecell{49.13\\{\footnotesize $\pm8.40$}} & 42.70\\
		Multi-modal (Ours) & \makecell{56.98\\{\footnotesize $\pm7.23$}} & \makecell{82.79\\{\footnotesize $\pm3.96$}} & \makecell{96.24\\{\footnotesize $\pm3.91$}} & \makecell{80.42\\{\footnotesize $\pm5.42$}} & \makecell{92.75\\{\footnotesize $\pm0.93$}} & \makecell{92.82\\{\footnotesize $\pm0.82$}} & \makecell{92.98\\{\footnotesize $\pm0.95$}} & \makecell{99.97\\{\footnotesize $\pm0.04$}} & \makecell{99.97\\{\footnotesize $\pm0.04$}} & 88.32\\
		\bottomrule
	\end{tabular*}
\end{table}

\subsection{Discussion}
\noindent\textbf{(1) Results for each fault type}

\begin{figure}
	\centering
	\subfigure[]{
		\begin{minipage}{0.23\linewidth}
			\centering
			\includegraphics[width=\linewidth]{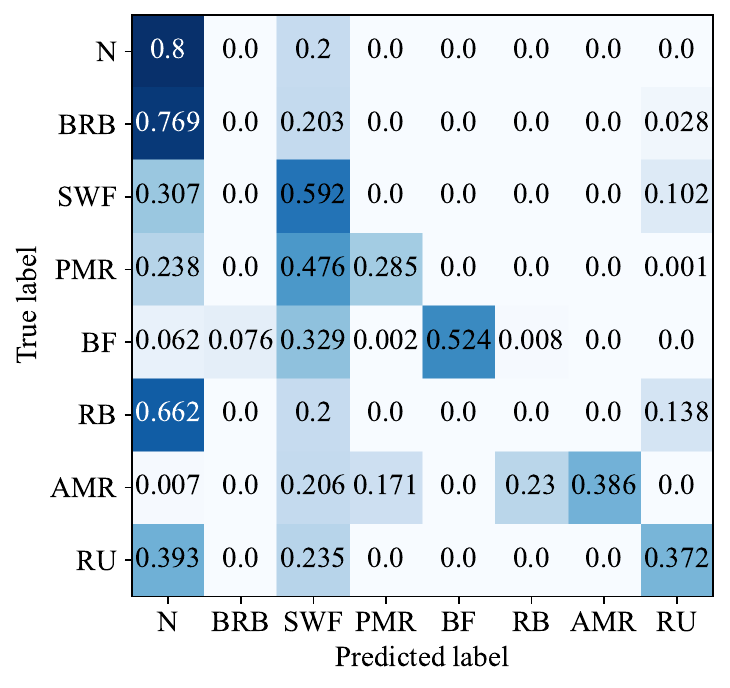}
		\end{minipage}
	}
	\subfigure[]{
		\begin{minipage}{0.23\linewidth}
			\centering
			\includegraphics[width=\linewidth]{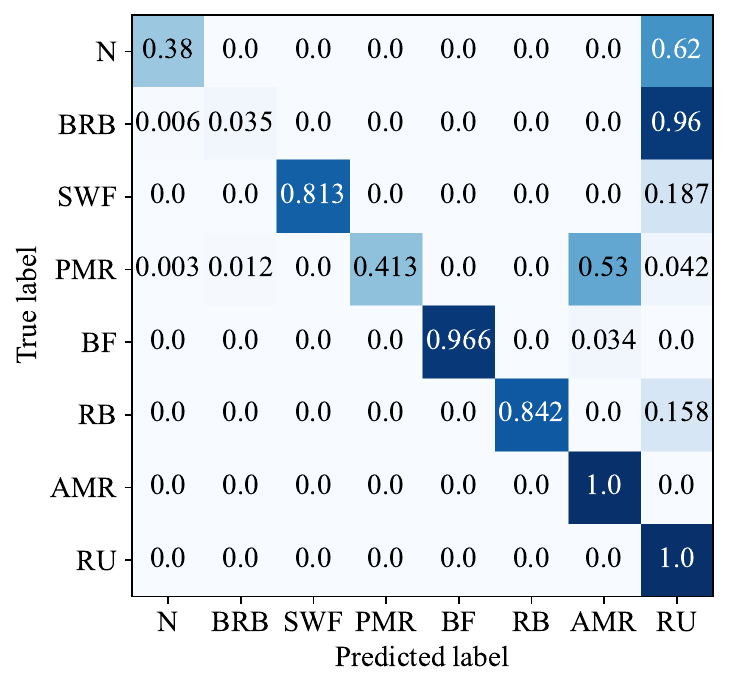}
		\end{minipage}
	}
	\subfigure[]{
		\begin{minipage}{0.23\linewidth}
			\centering
			\includegraphics[width=\linewidth]{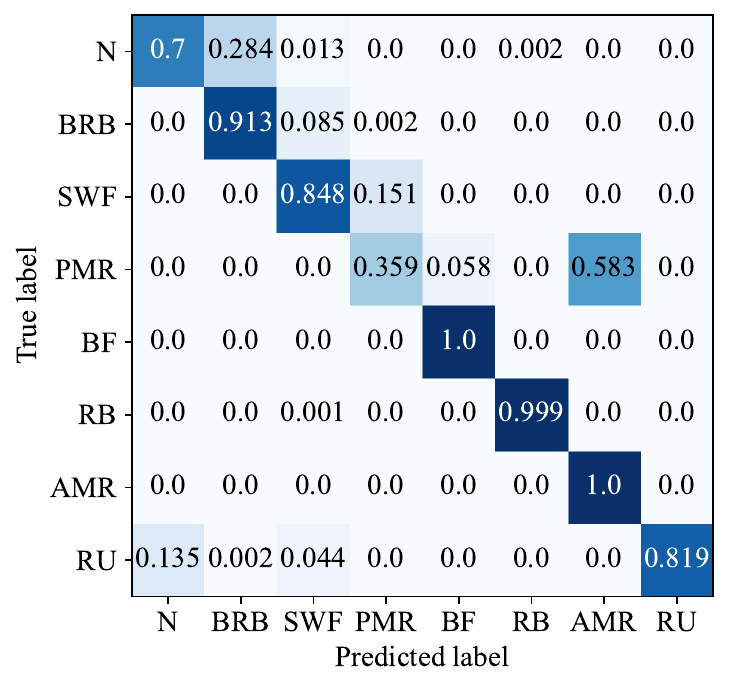}
		\end{minipage}
	}
	\subfigure[]{
		\begin{minipage}{0.23\linewidth}
			\centering
			\includegraphics[width=\linewidth]{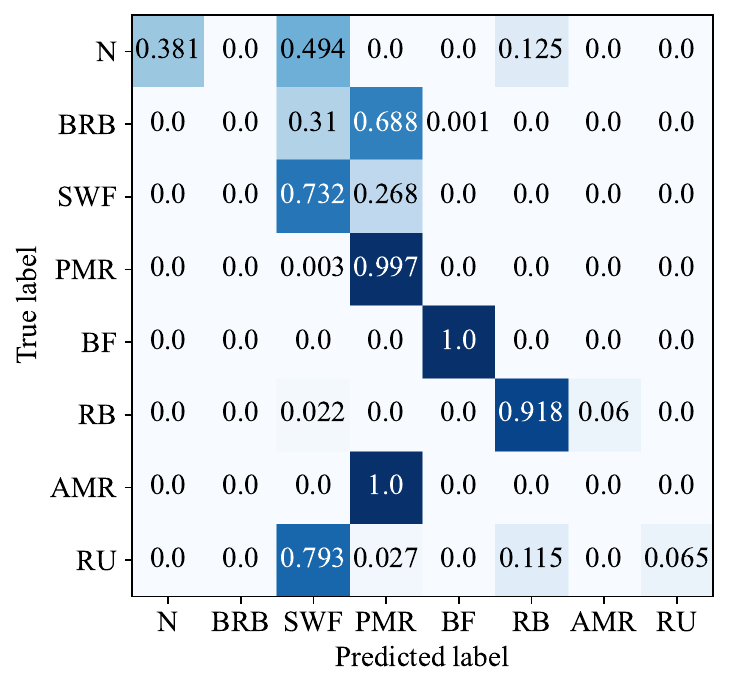}
		\end{minipage}
	}
	\subfigure[]{
		\begin{minipage}{0.23\linewidth}
			\centering
			\includegraphics[width=\linewidth]{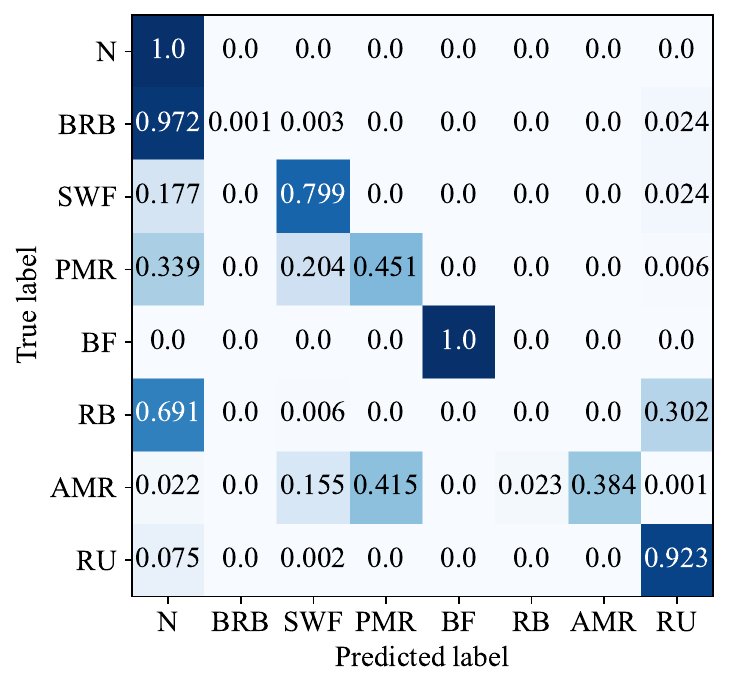}
		\end{minipage}
	}
	\subfigure[]{
		\begin{minipage}{0.23\linewidth}
			\centering
			\includegraphics[width=\linewidth]{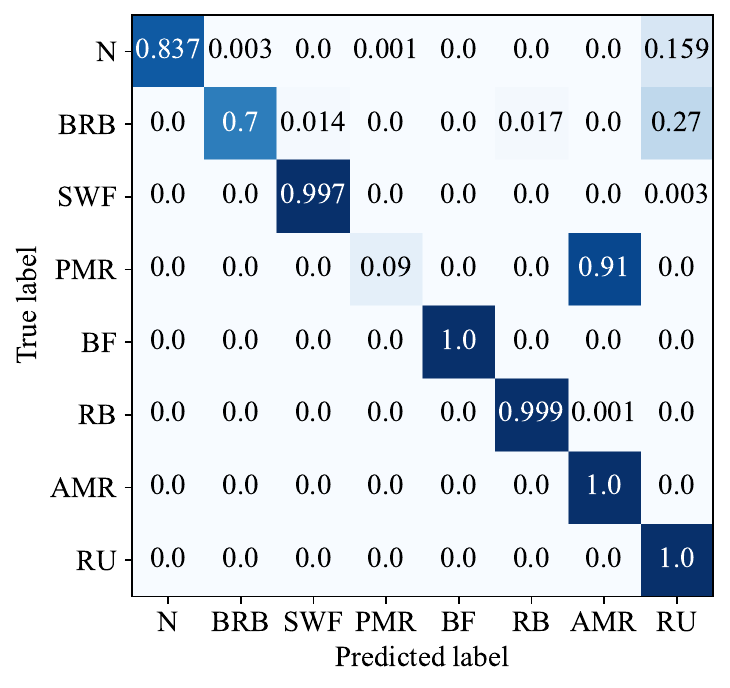}
		\end{minipage}
	}
	\subfigure[]{
		\begin{minipage}{0.23\linewidth}
			\centering
			\includegraphics[width=\linewidth]{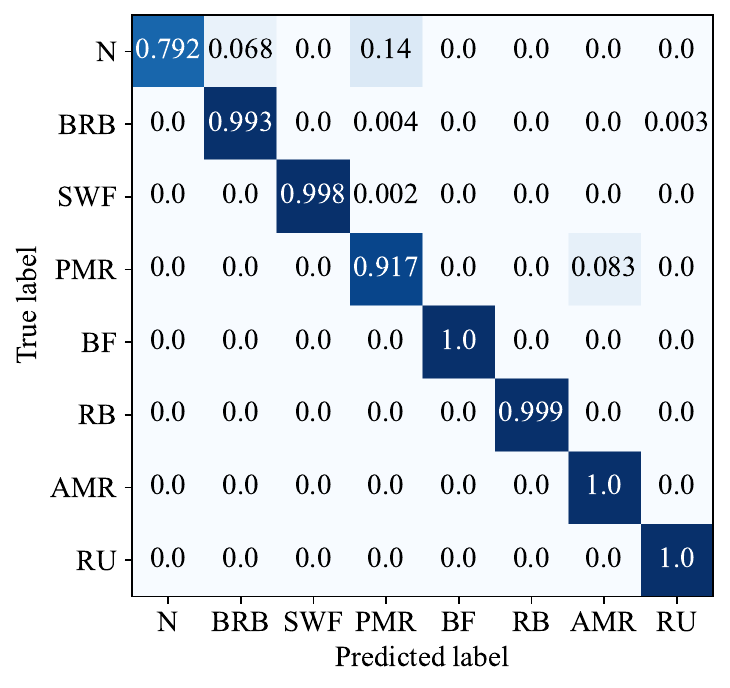}
		\end{minipage}
	}
	\subfigure[]{
		\begin{minipage}{0.23\linewidth}
			\centering
			\includegraphics[width=\linewidth]{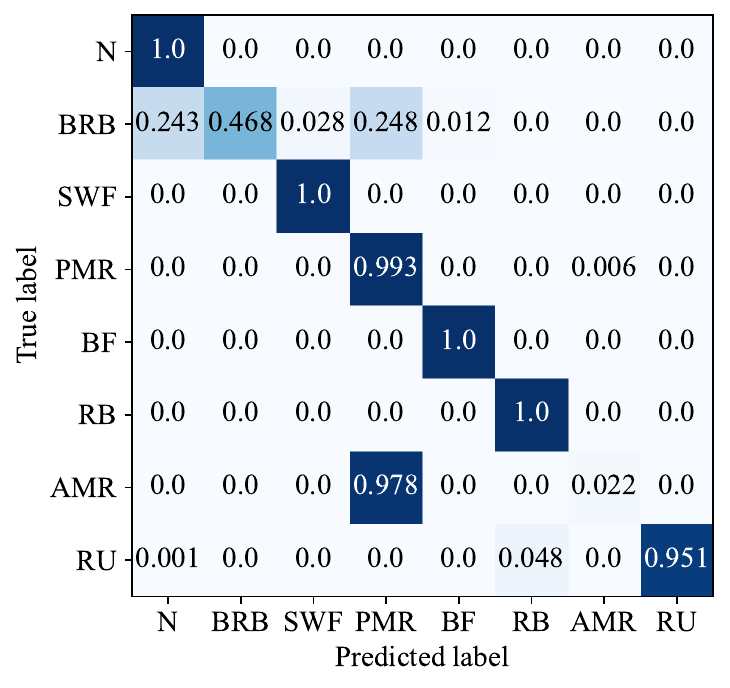}
		\end{minipage}
	}
	\caption{Confusion matrices in tasks T1-T4 of (a)-(d) \textit{Baseline} method and (e)-(h) the proposed method.}
	\label{fig:matrix}
\end{figure}

To further analyze the fault diagnosis performance, the confusion matrices of the \textit{Baseline} method and the proposed method are presented in Fig.~\ref{fig:matrix}. It can be observed that the \textit{Baseline} method exhibits noticeable misclassification among several fault categories. In particular, confusion frequently occurs between fault types of N and BRB, N and RB. This indicates that relying on simple multi-modal feature concatenation without effective disentanglement and adaptive fusion makes it difficult to distinguish faults from normal states under cross-condition settings. In contrast, the proposed method achieves substantially clearer diagonal patterns across all tasks, demonstrating improved class-wise discrimination. Most fault categories exhibit near-perfect classification accuracy without access to any samples in this domain, and most confusion observed in the \textit{Baseline} method is significantly alleviated. 

\noindent\textbf{(2) Hyperparameter sensitivity analysis}

\begin{figure}
	\centering
	\includegraphics[width=0.5\linewidth]{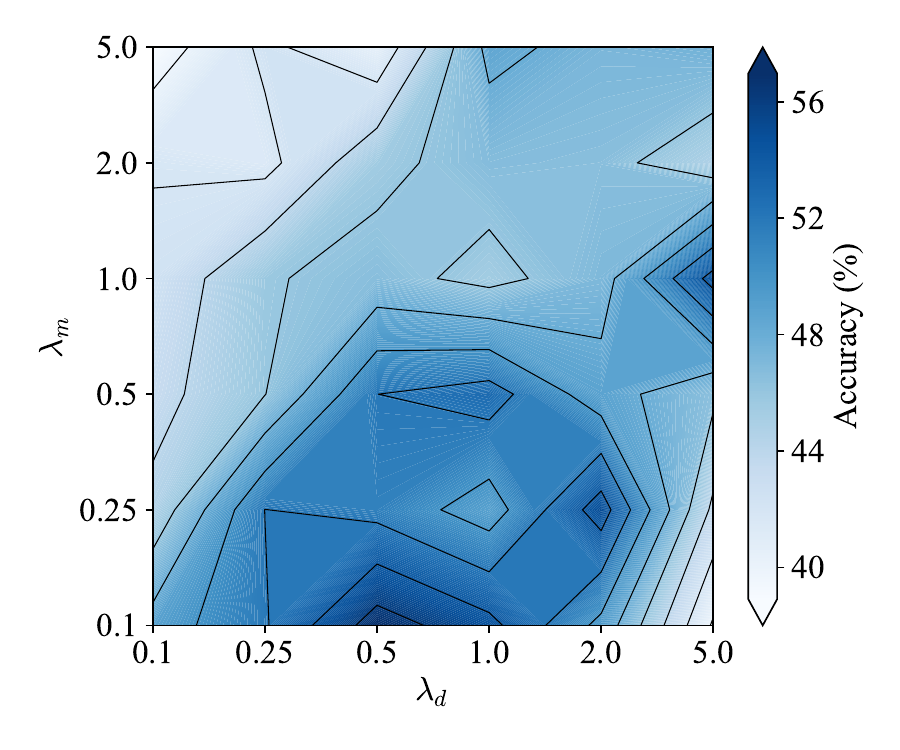}
	\caption{Fault diagnosis accuracy over different trade-off parameters.}
	\label{fig:para}
\end{figure}

Furthermore, model sensitivity on the two trade-off parameters $\lambda_m$ and $\lambda_d$ is analyzed on the first task, i.e., T1. The diagnosis accuracy over different parameter values is depicted in Fig.~\ref{fig:para}. It can be observed that the diagnosis accuracy varies smoothly with respect to both $\lambda_m$ and $\lambda_d$ within the middle-bottom region, indicating that the proposed method is not overly sensitive to the exact choice of trade-off parameters in specific ranges. Overall, relatively higher accuracies are achieved when $\lambda_m$ is set to small or moderate values, suggesting that excessive emphasis on modality-level disentanglement may suppress modality-specific discriminative information. In contrast, moderate values of $\lambda_d$ consistently contribute to performance improvement, highlighting the importance of domain-level disentanglement in reducing cross-domain discrepancies. Specifically, the best performance is obtained when $\lambda_m$ is around 0.1-0.25 and $\lambda_d$ lies in the range of 0.5-2.0, where a favorable balance between modality disentanglement and domain alignment is achieved. Consequently, $\lambda_m$ and $\lambda_d$ are set to 0.1 and 0.5, respectively, in this work.

\noindent\textbf{(3) Robustness to missing modalities}

In this paper, the proposed framework is primarily designed under a complete multi-modal setting, where vibration, current, and acoustic signals are jointly utilized for diagnostic model training and inference. However, complete and perfectly synchronized multi-modal measurements may not always be available during deployment in practical applications. To explicitly evaluate the robustness of the model to missing modalities in practical deployment, additional modality-missing experiments under unseen working conditions are designed. Specifically, the input of a specific modality is replaced with zeros to simulate the missing of this modality during inference to test model performance. The evaluation is performed on the same nine diagnosis tasks under unseen conditions as in the original experiments. It is worth noting that vibration signals typically play a dominant role in motor fault diagnosis, while current and acoustic signals provide complementary information. Therefore, the complete absence of vibration measurements is less common for reliable fault diagnosis in practical scenarios. As a result, we simulate a partial missing case by randomly masking one vibration channel for vibration missing scenario to reflect a sensor failure situation. It can be observed from the experiment results shown in Table~\ref{tab:modal_missing} that the model exhibits strong robustness to acoustic modality missing, with only marginal performance degradation observed across most tasks. When current modality is missing, the average diagnostic accuracy decreases by approximately 10\%, while a more noticeable decline is observed in the vibration-missing scenario, which is consistent with the relative importance of these modalities in motor fault diagnosis. Overall, the proposed framework maintains acceptable performance decline in most cases, demonstrating a certain degree of its inherent robustness to missing modalities during deployment.
\begin{table}[width=\linewidth,cols=11]
	\begin{threeparttable}
	\caption{Accuracy of modality missing cases on different tasks (\%).}\label{tab:modal_missing}
	\setlength{\tabcolsep}{8pt}
	\renewcommand{\arraystretch}{1.2}
	\begin{tabular*}{\tblwidth}{ccccccccccc}
		\toprule
		Missing Modality & T1 & T2 & T3 & T4 & T5 & T6 & T7 & T8 & T9 & Average\\
		\midrule
		No Missing & \makecell{56.98\\{\footnotesize $\pm7.23$}} & \makecell{82.79\\{\footnotesize $\pm3.96$}} & \makecell{96.24\\{\footnotesize $\pm3.91$}} & \makecell{80.42\\{\footnotesize $\pm5.42$}} & \makecell{92.75\\{\footnotesize $\pm0.93$}} & \makecell{92.82\\{\footnotesize $\pm0.82$}} & \makecell{92.98\\{\footnotesize $\pm0.95$}} & \makecell{99.97\\{\footnotesize $\pm0.04$}} & \makecell{99.97\\{\footnotesize $\pm0.04$}} & 88.32\\
		Vibration\tnote{1} & \makecell{38.61\\{\footnotesize $\pm6.78$}} & \makecell{67.69\\{\footnotesize $\pm5.14$}} & \makecell{75.10\\{\footnotesize $\pm3.03$}} & \makecell{63.26\\{\footnotesize $\pm5.41$}} & \makecell{69.90\\{\footnotesize $\pm3.16$}} & \makecell{69.94\\{\footnotesize $\pm3.89$}} & \makecell{70.16\\{\footnotesize $\pm3.11$}} & \makecell{74.79\\{\footnotesize $\pm6.25$}} & \makecell{75.68\\{\footnotesize $\pm6.32$}} & 67.23\\
		Current & \makecell{46.62\\{\footnotesize $\pm5.30$}} & \makecell{61.47\\{\footnotesize $\pm14.94$}} & \makecell{85.32\\{\footnotesize $\pm4.94$}} & \makecell{69.23\\{\footnotesize $\pm6.91$}} & \makecell{83.98\\{\footnotesize $\pm3.16$}} & \makecell{84.04\\{\footnotesize $\pm3.20$}} & \makecell{84.44\\{\footnotesize $\pm3.23$}} & \makecell{93.51\\{\footnotesize $\pm4.99$}} & \makecell{93.51\\{\footnotesize $\pm4.99$}} & 78.01\\
		Acoustic & \makecell{50.83\\{\footnotesize $\pm6.89$}} & \makecell{82.43\\{\footnotesize $\pm3.70$}} & \makecell{96.05\\{\footnotesize $\pm3.83$}} & \makecell{80.33\\{\footnotesize $\pm6.02$}} & \makecell{92.37\\{\footnotesize $\pm0.75$}} & \makecell{92.38\\{\footnotesize $\pm0.82$}} & \makecell{92.70\\{\footnotesize $\pm0.81$}} & \makecell{99.97\\{\footnotesize $\pm0.04$}} & \makecell{99.97\\{\footnotesize $\pm0.04$}} & 87.45\\
		\bottomrule
	\end{tabular*}
	\begin{tablenotes}
		\item [1] Channel randomly missing.
	\end{tablenotes}
	\end{threeparttable}
\end{table}

\noindent\textbf{(4) Feature disentanglement analysis}

\begin{figure}
	\centering
	\subfigure[]{
		\begin{minipage}{0.30\linewidth}
			\centering
			\includegraphics[width=\linewidth]{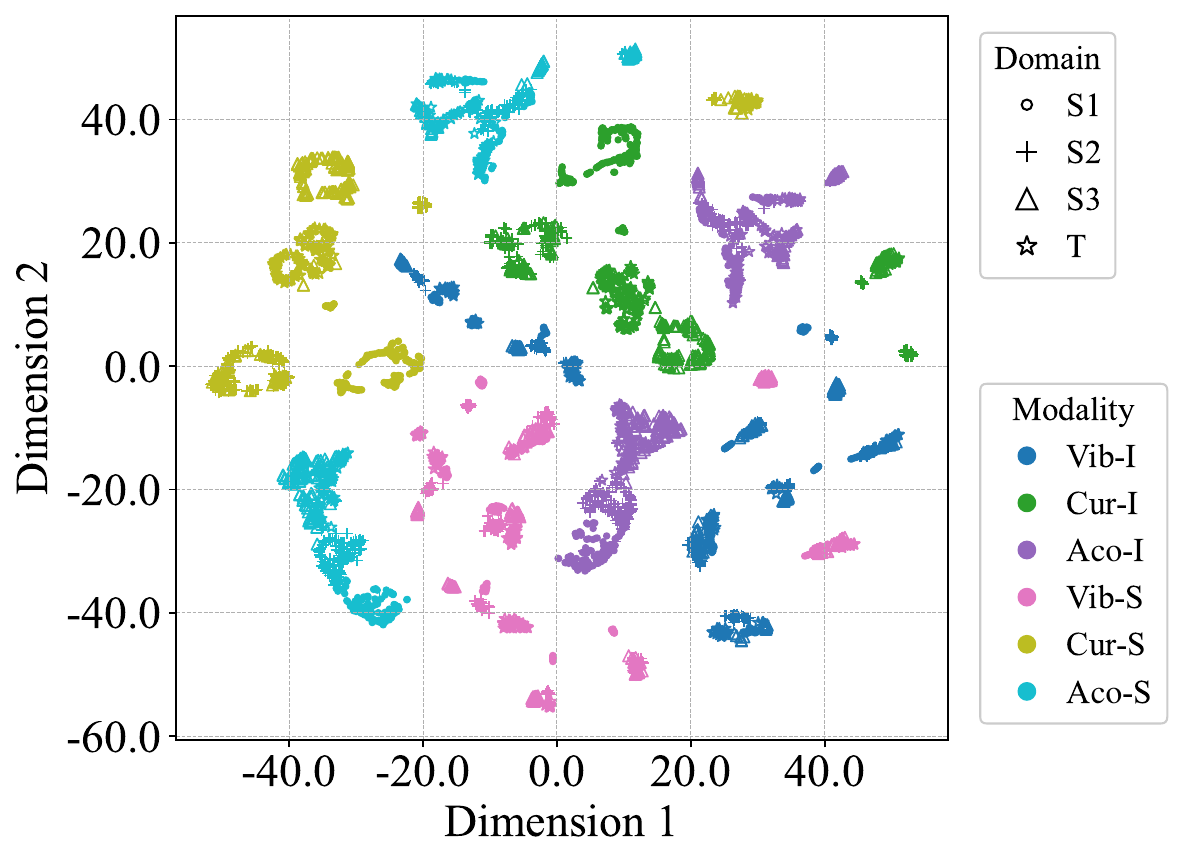}
		\end{minipage}
	}
	\subfigure[]{
		\begin{minipage}{0.30\linewidth}
			\centering
			\includegraphics[width=\linewidth]{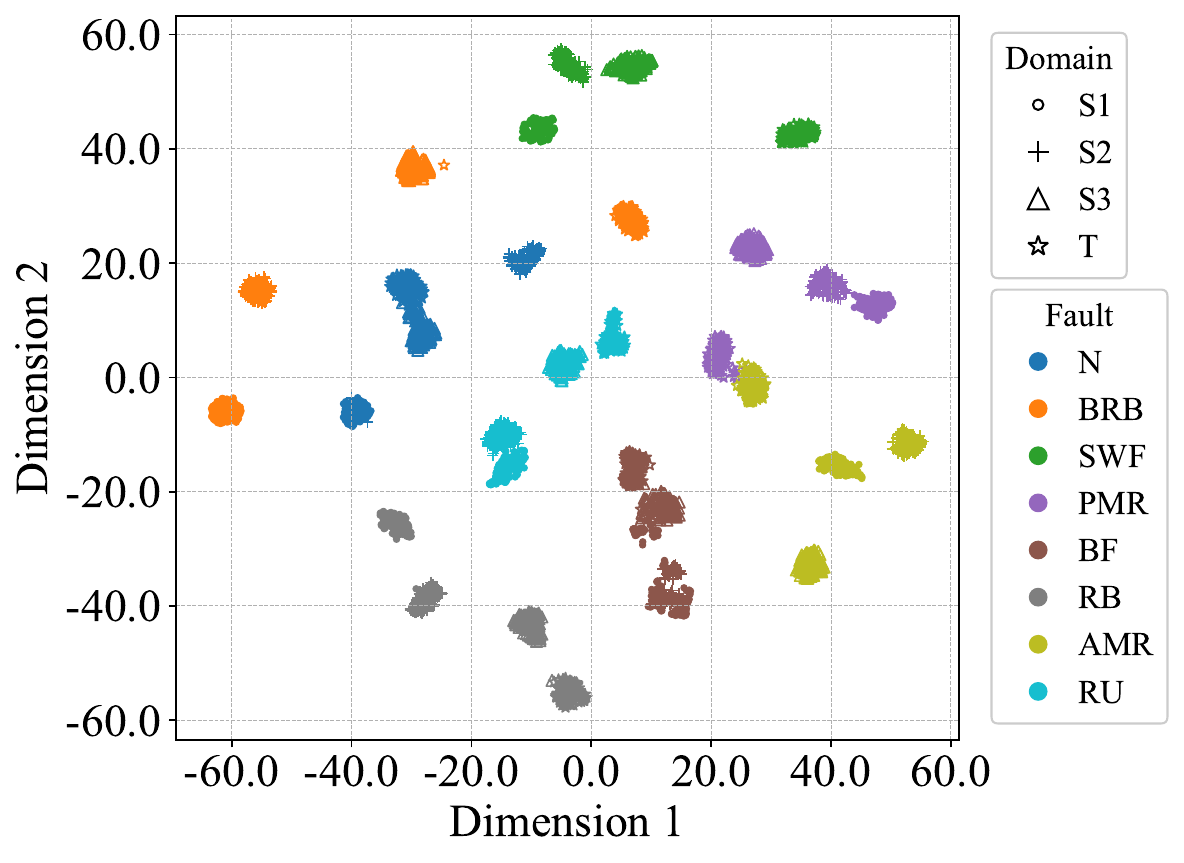}
		\end{minipage}
	}
	\subfigure[]{
		\begin{minipage}{0.30\linewidth}
			\centering
			\includegraphics[width=\linewidth]{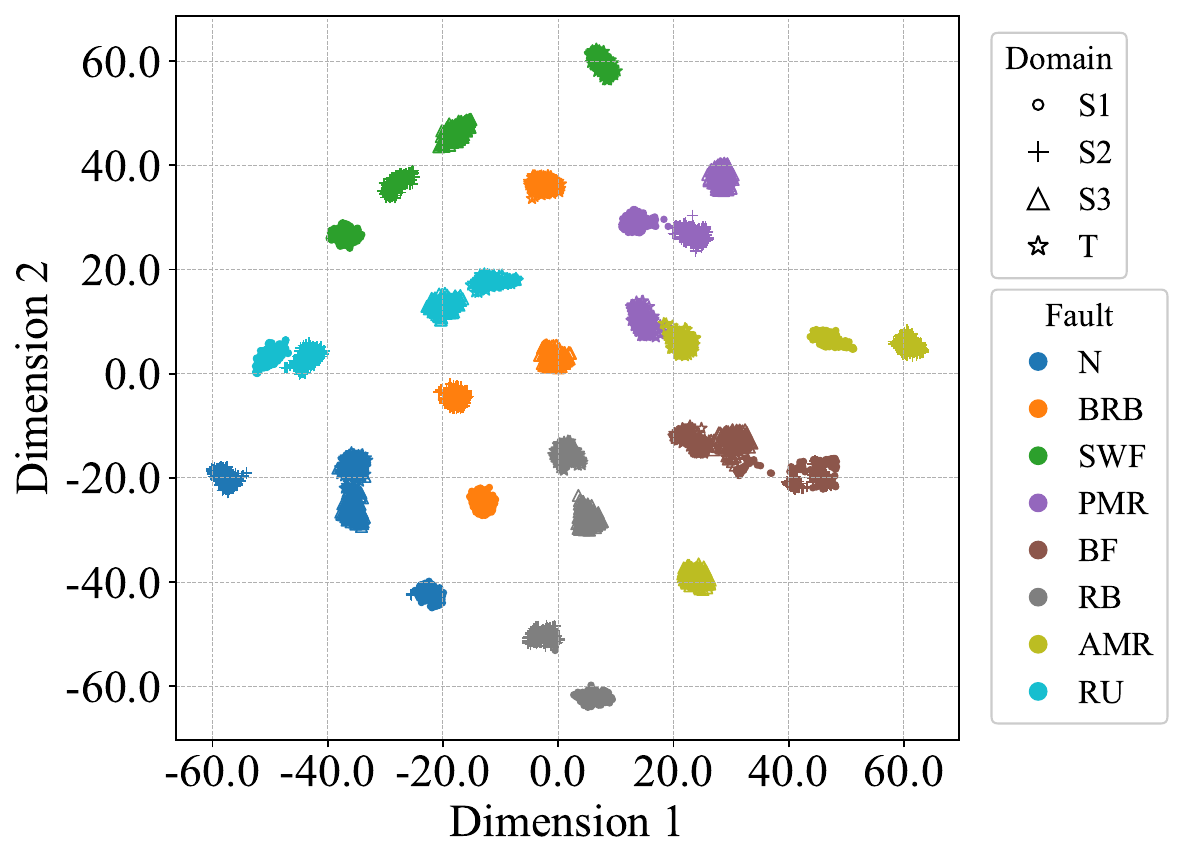}
		\end{minipage}
	}
	\subfigure[]{
		\begin{minipage}{0.30\linewidth}
			\centering
			\includegraphics[width=\linewidth]{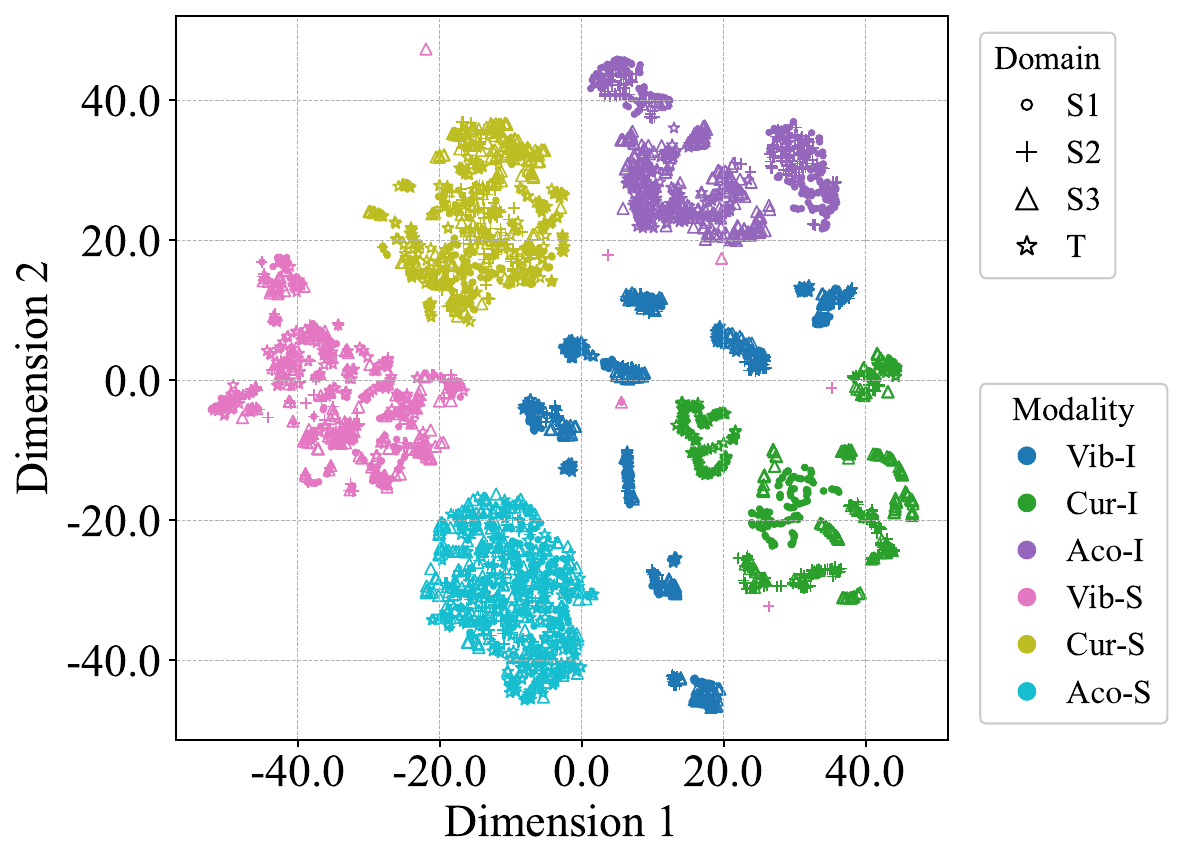}
		\end{minipage}
	}
	\subfigure[]{
		\begin{minipage}{0.30\linewidth}
			\centering
			\includegraphics[width=\linewidth]{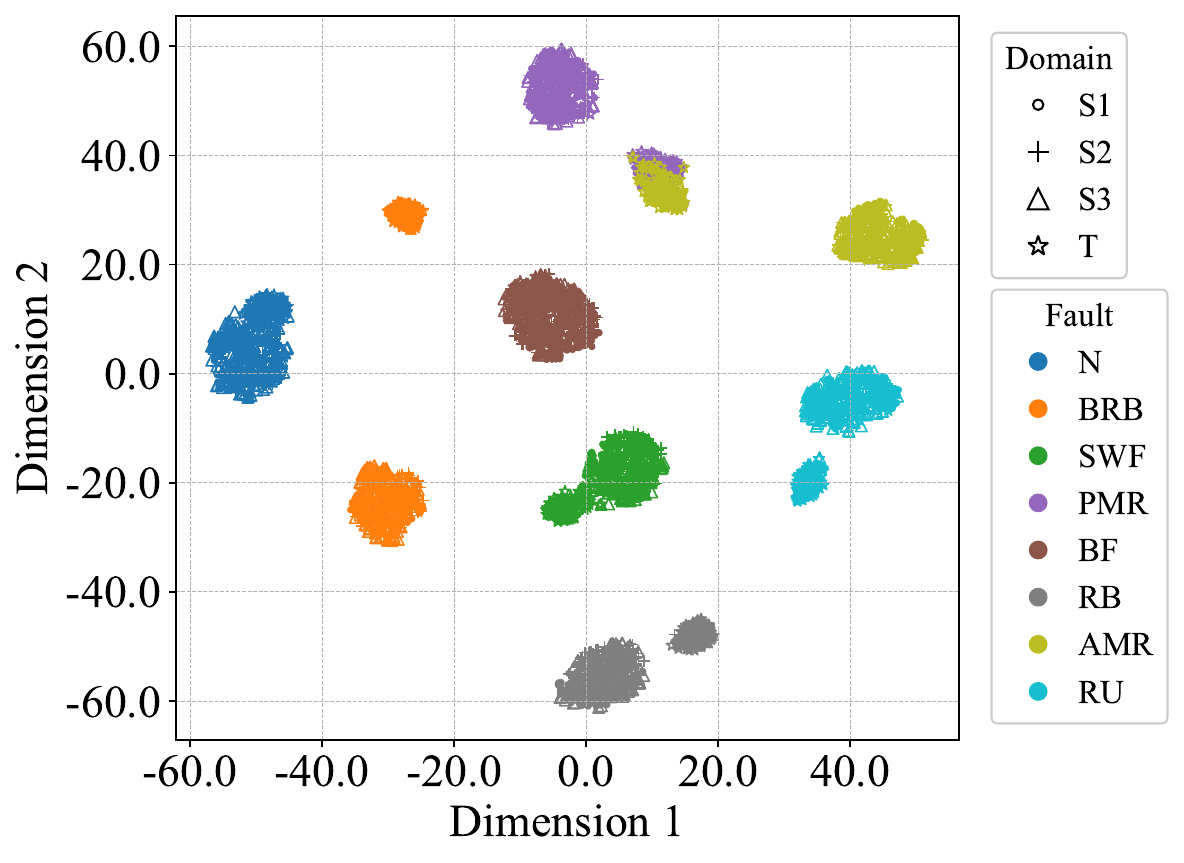}
		\end{minipage}
	}
	\subfigure[]{
		\begin{minipage}{0.30\linewidth}
			\centering
			\includegraphics[width=\linewidth]{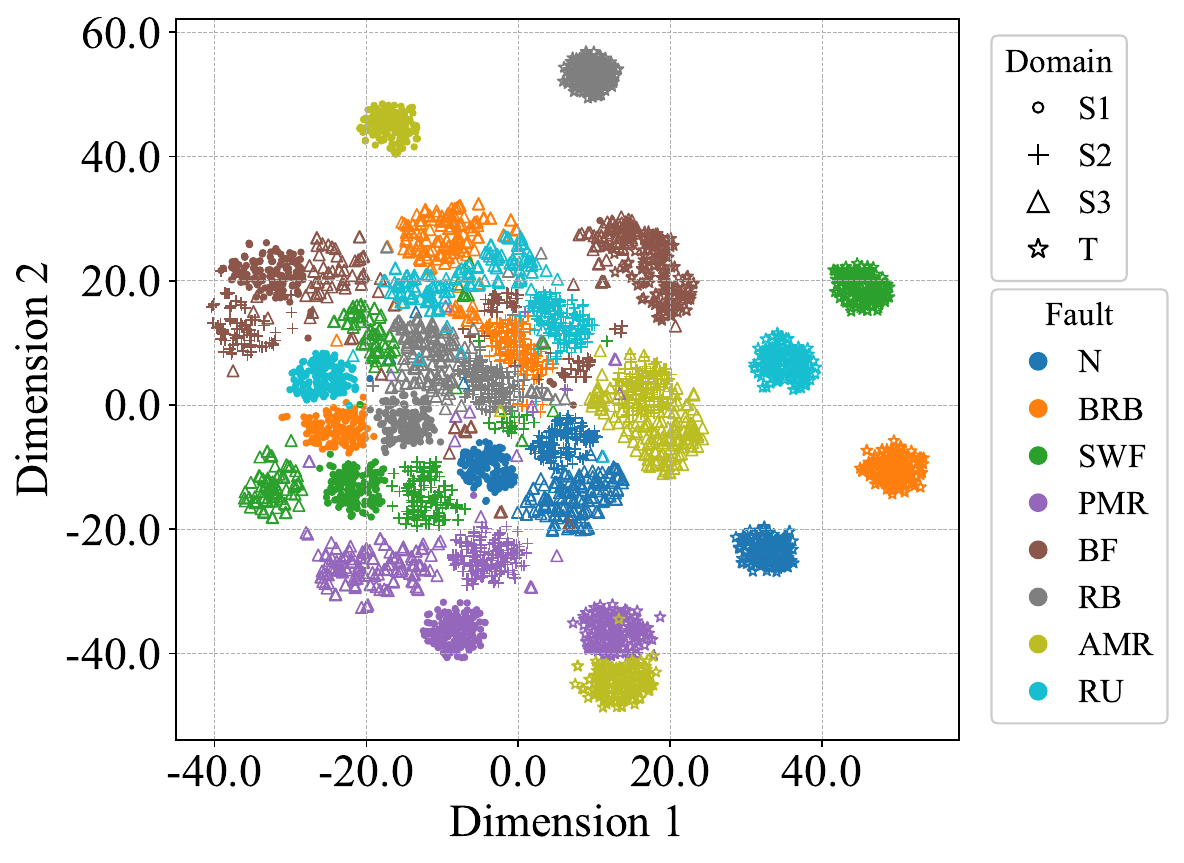}
		\end{minipage}
	}
	\caption{Feature visualization with t-SNE: modality-invariant/modality-specific features without (a) and with dual-disentanglement (d), domain-invariant features without (b) and with dual-disentanglement (e), and domain-specific features without (c) and with dual-disentanglement (f). Vib, Cur, and Aco denote vibration, current, and acoustic modalities, respectively; I and S denote modality-invariant and modality-specific features, respectively. S1, S2, S3, and T denote the three source domains and the target domain, respectively.}
	\label{fig:visualization}
\end{figure}

To evaluation the effectiveness of the proposed dual-disentanglement framework, the learned representations are analyzed from both qualitative and quantitative perspectives. First, to qualitatively demonstrate the disentangled feature distributions, t-SNE (t-distributed Stochastic Neighbor Embedding)~\cite{t-sne} is used to visualize the learned modality-level and domain-level representations, which is trained on the first three working conditions and test on the fourth condition (Task T4). Features from models without and with dual-disentanglement are visualized for comparison in Fig.~\ref{fig:visualization}.

For modality-level disentanglement visualized in the first column, the model without dual disentanglement produces relatively mixed and scattered invariant and specific representations, indicating that shared and modality-dependent information is not clearly separated. In contrast, the proposed method yields a more structured distribution with more compact and structured clusters. The modality-specific features become more distinguishable across different sensing modalities, while the overlap between invariant and specific representations is reduced. The modality-invariant features of different modalities are also drawn closer in the embedding space, especially for vibration and current modalities, which exhibit a certain degree of mixing without clear cluster boundaries. It should be noted that these invariant representations are not expected to become completely indistinguishable across modalities, because the proposed method adopts MMD-based distribution alignment rather than strong adversarial modality confusion and the modality-level disentanglement is jointly optimized with the primary diagnostic objective with a small trade-off coefficient. Overall, the visualization indicates that the proposed framework promotes a more structured separation between shared and modality-specific information.

For domain-level representations, the proposed method also shows clearer effects. For the domain-invariant features shown in the middle column, the representations learned by the proposed method form more compact and discriminative fault clusters compared with the model without dual disentanglement. Samples from different source domains and the unseen target domain tend to be mapped into the same fault-specific regions, suggesting that the learned domain-invariant representations reduce the influence of working-condition variations. In contrast, the domain-specific representations illustrated in the last column exhibit stronger dispersion across different working conditions and are less compact in terms of fault-category clustering. Samples from different domains tend to occupy different local regions, which suggests that the domain-specific branch preserves working-condition-related characteristics. Meanwhile, the fault labels in Fig.~\ref{fig:visualization}(f) are not completely random or indistinguishable, indicating that the domain-specific representations still retain partial fault-related information. This observation is consistent with the expected role of domain-specific representations, which are designed to preserve condition-dependent characteristics as complementary information.

Furthermore, the cross-covariance between invariant and specific representations is calculated to quantitatively evaluate the disentanglement degree between the two subspaces. As shown in Table~\ref{tab:cov}, relatively large cross-covariance values are observed at both modality and domain levels without disentanglement, indicating that invariant and specific features remain strongly correlated. After introducing the proposed dual-disentanglement, the average modality-level cross-covariance decreases from 0.0923 to 0.0003, while the average domain-level cross-covariance decreases from 1.5717 to 0.0010. Notably, consistently low covariance is obtained for the unseen target domain, indicating that the disentanglement is not only effectively trained on the source domain, but also able to be generalized to disentangle features on the completely unseen domain. These significant reductions demonstrate that the proposed disentanglement effectively suppress redundancy and promotes the decoupling between invariant and specific representations.
\begin{table}[width=\linewidth,cols=10]
	\caption{Average cross-covariance between invariant and specific features}\label{tab:cov}
	\setlength{\tabcolsep}{7pt}
	\renewcommand{\arraystretch}{1.2}
	\begin{tabular*}{\tblwidth}{c|cccc|ccccc}
		\toprule
		\multirow{2}{*}{Method} & \multicolumn{4}{c|}{Modality-level} & \multicolumn{5}{c}{Domain-level}\\
		\cline{2-10} & Vibration & Current & Acoustic & Average & Source 1 & Source 2 & Source 3 & Target & Average\\
		\midrule
		w/o dis & 0.1498 & 0.0400 & 0.0871 & 0.0923 & 1.6283 & 1.8321 & 1.4966 & 1.3297 & 1.5717\\
		w/ dis (Ours) & 0.0004 & 0.0002 & 0.0002 & 0.0003 & 0.0010 & 0.0008 & 0.0009 & 0.0014 & 0.0010\\
		\bottomrule
	\end{tabular*}
\end{table}

\begin{figure}
	\centering
	\subfigure[]{
		\begin{minipage}{0.23\linewidth}
			\centering
			\includegraphics[width=\linewidth]{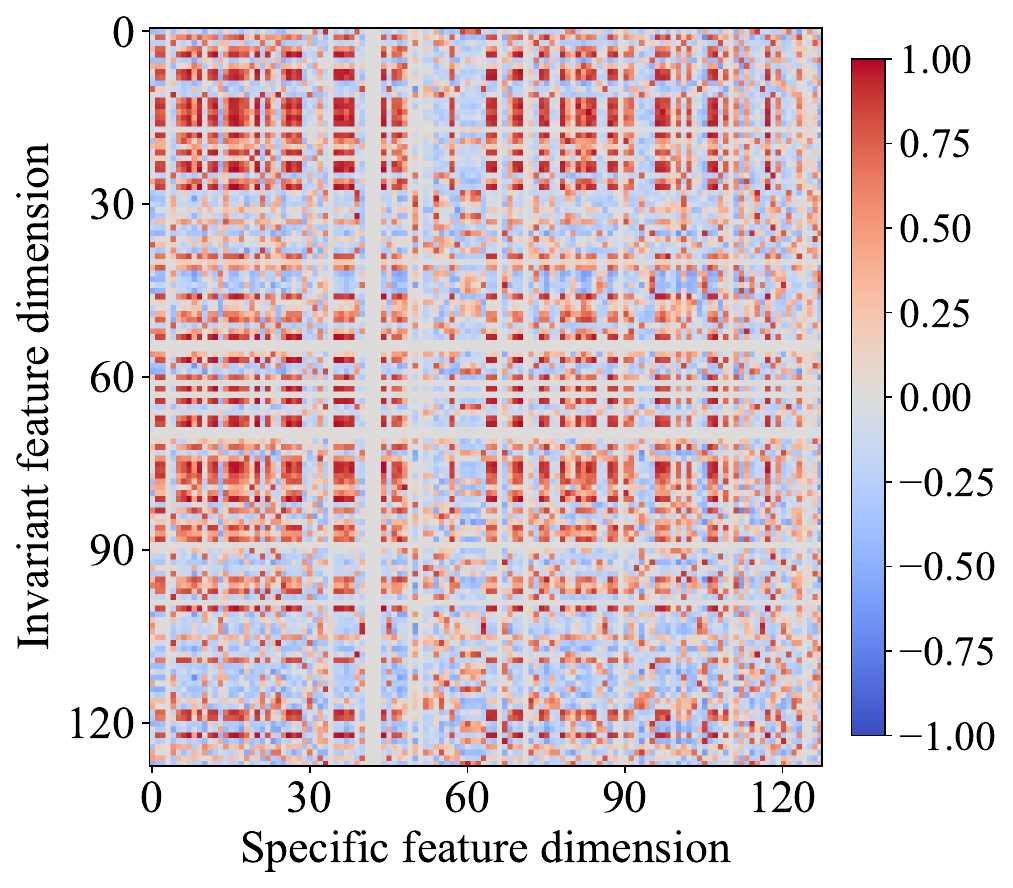}
		\end{minipage}
	}
	\subfigure[]{
		\begin{minipage}{0.23\linewidth}
			\centering
			\includegraphics[width=\linewidth]{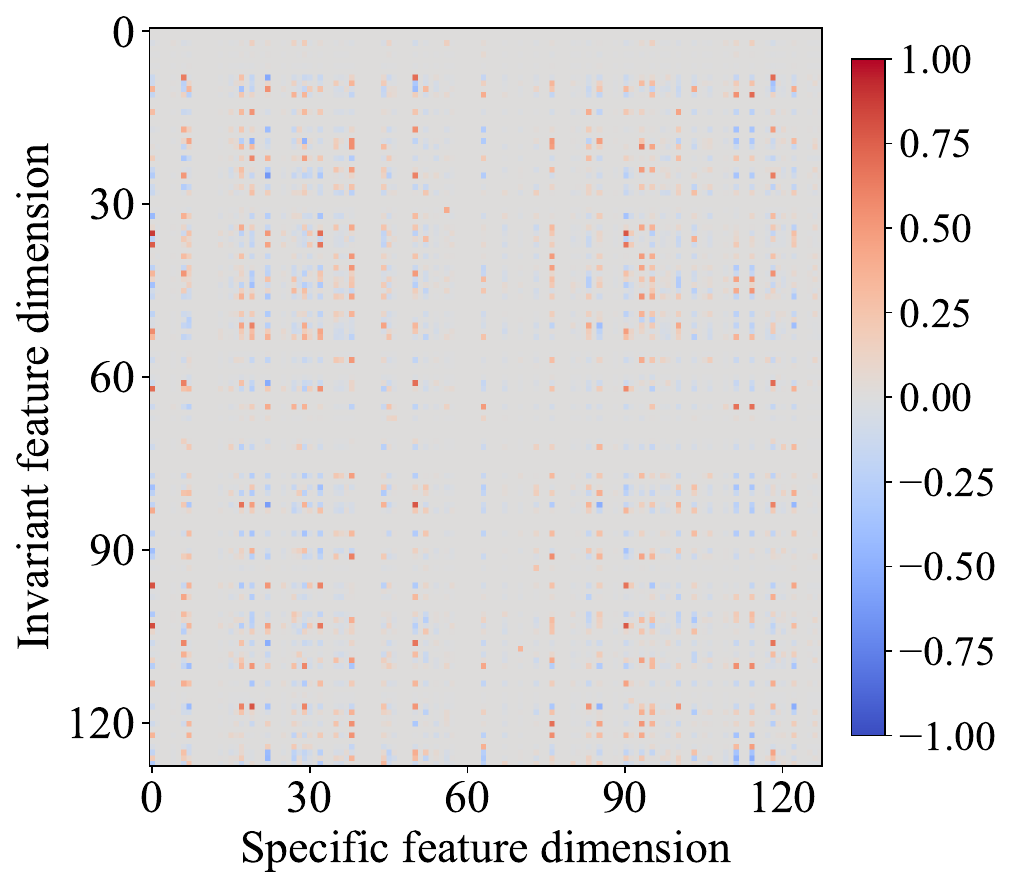}
		\end{minipage}
	}
	\subfigure[]{
		\begin{minipage}{0.23\linewidth}
			\centering
			\includegraphics[width=\linewidth]{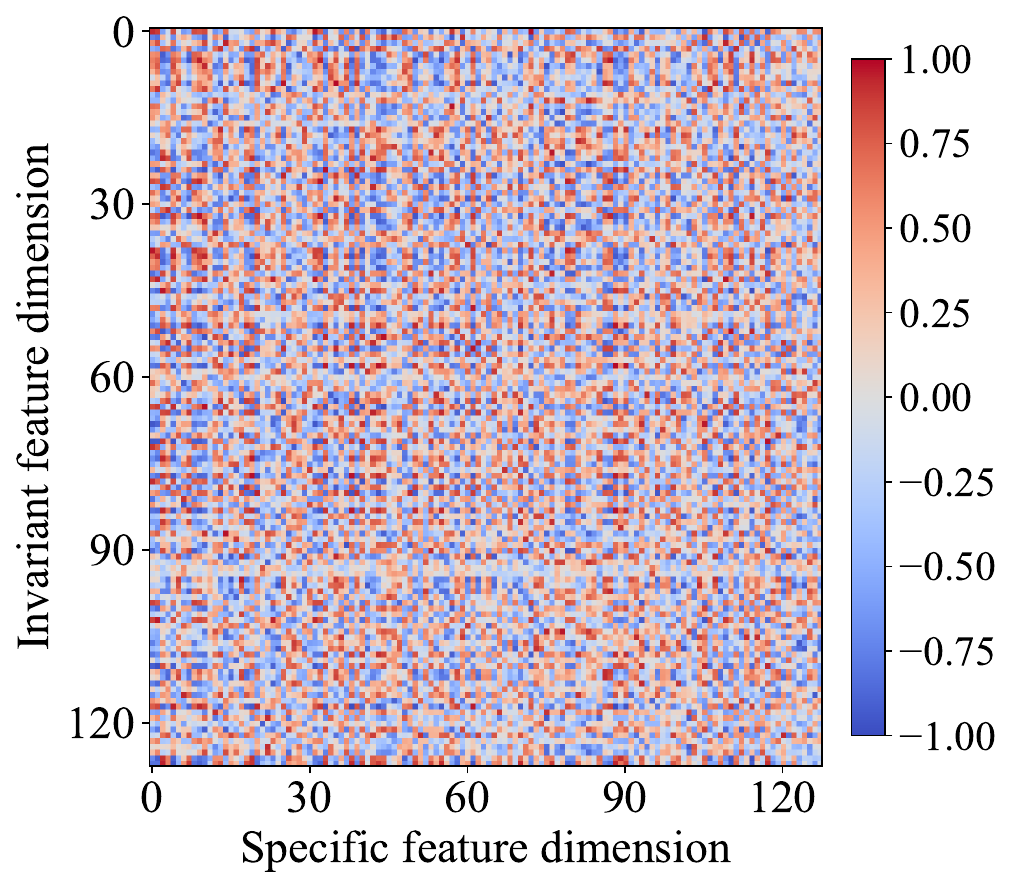}
		\end{minipage}
	}
	\subfigure[]{
		\begin{minipage}{0.23\linewidth}
			\centering
			\includegraphics[width=\linewidth]{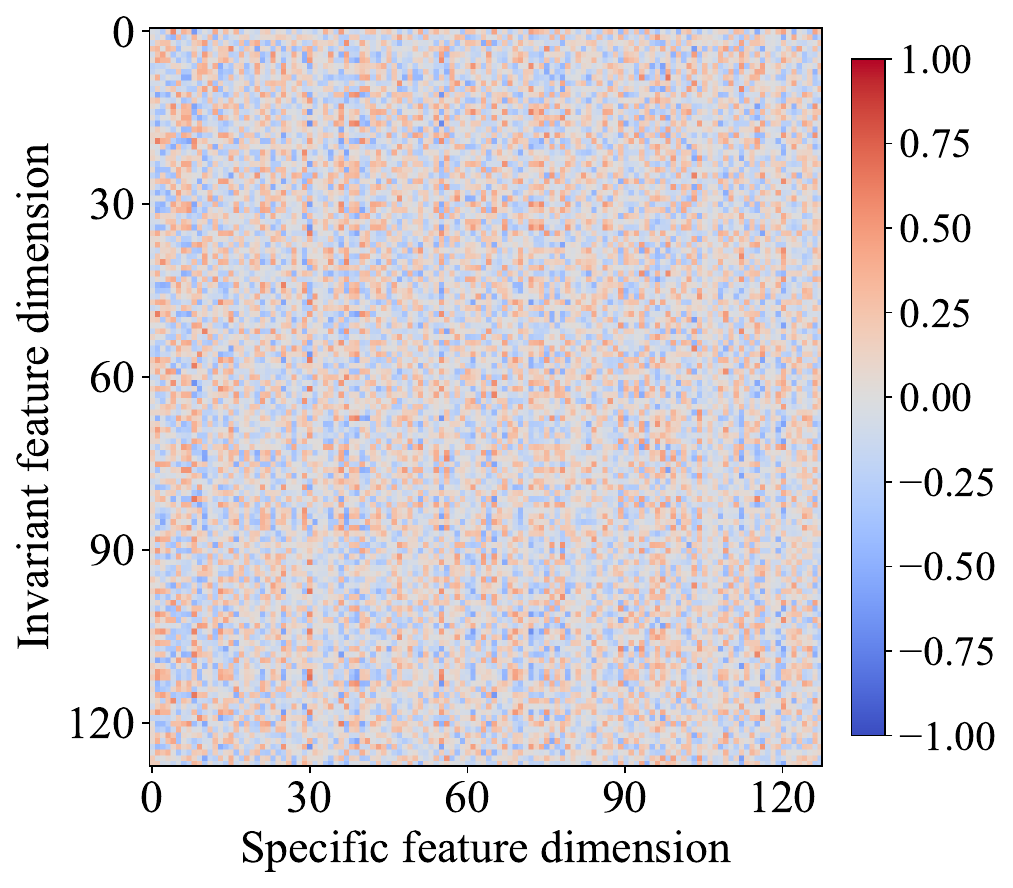}
		\end{minipage}
	}
	\caption{Cross-correlation heatmaps: between modality-invariant and modality-specific features without (a) and with dual-disentanglement (b) of vibration modality, and between domain-invariant and domain-specific features without (c) and with dual-disentanglement (d) of target domain.}
	\label{fig:correlation}
\end{figure}

To further illustrate the dependence between invariant and specific features, Pearson cross-correlation heatmaps between two types of representations are calculated. Fig.~\ref{fig:correlation} depicts the modality-level correlation for the vibration modality and the domain-level correlation for the target domain, while similar phenomena can be observed for other modalities and domains. Without dual disentanglement, dense positive and negative correlation responses can be observed, indicating strong dependence between the two feature subspaces. By contrast, the proposed method produces much weaker correlation responses, and most elements in the heatmaps are close to zero. This trend is consistent with the cross-covariance results and confirms that the proposed framework effectively decouples invariant and specific representations.

Overall, the above analysis consistently demonstrates that the proposed dual-disentanglement framework promotes the separation of shared and specific information at both modality and domain levels. Such a structured representation is beneficial for reducing redundant information, preserving complementary characteristics, and improving fault diagnosis under unseen working conditions.

\noindent\textbf{(5) Computational complexity analysis}

Computational complexity of a fault diagnosis model is crucial to practical feasibility. The proposed model consists of three encoders, a feature embedding module, a cross-attention-based fusion module, and a final classifier. Taking the notations in Section~\ref{sec_overview}, the computational cost of the three encoders can be expressed as
\begin{equation}
	\mathcal{O}_{enc} = \sum_{l\in L_v}\mathcal{O}(H_v^{(l)} W_v^{(l)} C_{v,in}^{(l)} C_{v,out}^{(l)} k_l^2 ) + \sum_{l\in L_c}\mathcal{O}(L_c^{(l)} C_{c,in}^{(l)} C_{c,out}^{(l)} k_l) + \sum_{l\in L_a}\mathcal{O}(H_a^{(l)} W_a^{(l)} C_{a,in}^{(l)} C_{a,out}^{(l)} k_l^2 ),
\end{equation}
where $L_v$, $L_a$, and $L_c$ denote the convolutional layers in each encoder, and $k_l$ is the convolutional kernel size. After feature extraction, the embedding layer results in a cost of $\mathcal{O}_{emb}=\mathcal{O}(3d^2)$, where $d$ is the feature embedding dimension. The complexity of multi-modal fusion module and the subsequent embedding layer is dominated by linear projections, i.e., $\mathcal{O}_{fusion} =\mathcal{O}(d^2)$. Finally, the classifier consists of a linear layer mapping from $d$ to $K$ classes, yielding $\mathcal{O}_{cls} =\mathcal{O}(dK)$. The overall inference complexity of the proposed framework can be summarized as
\begin{equation}
	\mathcal{O}_{infer} =\mathcal{O}_{enc} +\mathcal{O}(d^2)+\mathcal{O}(dK).
\end{equation}
Consequently, the total computational cost is dominated by the three encoders. The additional overhead introduced by the embedding, fusion, and classification modules is therefore marginal.

During training, cross-domain mixed fusion and dual-level disentanglement objectives are introduced. The cross-domain mixed fusion strategy performs sample-level linear combination, introducing a cost of $\mathcal{O}_{mix} =\mathcal{O}(B\cdot N_e)$, where $B$ is the batch size and $N_e$ is the number of elements per sample. Obviously, its cost is negligible, i.e., $\mathcal{O}_{mix}\ll \mathcal{O}_{enc} $. The additional computational cost incurred by dual disentanglement objectives can be approximated as
\begin{equation}
	\mathcal{O}_{dis}=\mathcal{O}(P_{MMD}B^2d)+\mathcal{O}(P_{cov}Bd^2),
\end{equation}
where $P_{MMD}$  and $P_{cov}$ denote the number of pairwise alignment and covariance computations, respectively. The additional computational cost of $\mathcal{O}_{mix}$ and $\mathcal{O}_{dis}$ is only required during training and is relatively small compared to that of three encoders, which is still the dominant cost in the training stage.

A detailed computational complexity evaluation is also included with some key metrics, including the total number of trainable parameters, Floating Point Operations per Second (FLOPs), the training time per epoch, and the inference time per sample. These metrics of some key ablation variants are reported as well in Table~\ref{tab:computational}, which are all tested under the same experiment setting. Notably, the first three variants represent the single-modal models which exploit only one modality input with one encoder. It can be observed that the proposed multi-modal framework inevitably introduces higher computational cost compared to single-modal models. Nevertheless, the three single-modal models exhibit limited representation capacity and performance. Compared with the \textit{baseline} multi-modal model, the proposed method increases computational complexity, while the increase is moderate especially for the inference time. It is noteworthy that variants such as \textit{w/o modality-dis}, \textit{w/o domain-dis}, and \textit{w/o dis} exhibit very similar FLOPs and inference time compared to the full model. This is consistent with the theoretical analysis, as the disentanglement mechanisms mainly introduce additional loss computations during training. The \textit{w/o mix} variant shows nearly identical inference complexity to the full model, while achieving a shorter training time due to absence of the cross-domain mixed fusion operation. Finally, the \textit{concat} variant does not significantly reduce the overall complexity, as the dominant cost still comes from the encoders.

Overall, the results demonstrate that the proposed framework achieves a favorable trade-off between performance and computational efficiency. Although it introduces moderate additional parameters and training costs, the increase in inference complexity is relatively small. The inference latency is approximately 3.7 ms per sample, which is sufficiently low for condition monitoring tasks, demonstrating that the model is suitable for practical deployment in industrial fault diagnosis scenarios.
\begin{table}[width=.75\linewidth,cols=5]
	\caption{Computational complexity metrics of the proposed model and different variants.}\label{tab:computational}
	\setlength{\tabcolsep}{12pt}
	\renewcommand{\arraystretch}{1.2}
	\begin{tabular*}{\tblwidth}{ccccc}
		\toprule
		Method & \makecell{Training time\\per epoch (s)} & \makecell{Trainable\\parameters} & FLOPs & \makecell{Inference time\\per sample (s)}\\
		\midrule
		Vibration-only & 0.92 & 1.29 M & 8.74 M & 0.0008\\
		Current-only & 0.83 & 0.51 M & 4.08 M & 0.0009\\
		Acoustic-only & 0.85 & 1.29 M & 19.82 M & 0.0009\\\hline
		Baseline & 1.30 & 2.88 M & 32.44 M & 0.0026\\
		w/o dis & 2.15 & 3.69 M & 33.03 M & 0.0035\\
		w/o modality-dis & 2.20 & 3.88 M & 33.23 M & 0.0036\\
		w/o domain-dis & 2.21 & 3.89 M & 33.23 M & 0.0035\\
		w/o mix & 1.46 & 4.08 M & 33.42 M & 0.0037\\
		concat & 2.07 & 3.09 M & 32.63 M & 0.0027\\
		Ours & 2.24 & 4.08 M & 33.42 M & 0.0037\\
		\bottomrule
	\end{tabular*}
\end{table}

\section{Conclusion}\label{sec_conclusion}
In this work, a multi-modal cross-domain mixed fusion model with dual disentanglement is proposed for fault diagnosis under unseen working conditions. A dual-level disentanglement strategy is introduced to decouple modality-invariant and modality-specific features, as well as domain-invariant and domain-specific representations, enabling generalizable feature learning across operating conditions. A multi-modal cross-domain mixed fusion strategy is proposed to augment modality and domain diversity for enhanced feature robustness. Furthermore, a triple-modal fusion mechanism is designed to adaptively integrate multi-modal representations for comprehensive information integration. Extensive experiments conducted under both constant and time-varying working conditions demonstrate that the proposed method consistently outperforms state-of-the-art domain generalization and multi-modal fusion approaches. Ablation studies further verify the effectiveness of each proposed component and the superiority of multi-modal heterogeneous information fusion. In future work, we plan to extend the proposed framework to more complex industrial scenarios involving more realistic operating conditions, as well as investigate its applicability to other rotating machinery. In addition, we will further investigate more flexible multi-modal learning strategies that can explicitly handle incomplete-modality scenarios by incorporating modality dropout or modality masking mechanisms during training to improve robustness against unexpected modality missing.

\section*{Acknowledgements}
This research was supported by the National Natural Science Foundation of China (No. 52405122 and No. 52375109), the National Key Research and Development Program of China (No. 2023YFB3408502), and the China Postdoctoral Science Foundation (No. 2025M771378).

\bibliographystyle{model1-num-names}

\bibliography{cas-refs}

\vskip3pt

\end{document}